%% file: main.tex
\documentclass[10pt,twocolumn,letterpaper]{article}

\usepackage{iccv}
\usepackage{times}
\usepackage{epsfig}
\usepackage{graphicx}
\usepackage{amsmath}
\usepackage{amssymb}
\usepackage{booktabs}
\usepackage{tabularx}
\usepackage[accsupp]{axessibility}

\usepackage[breaklinks=true,bookmarks=false,colorlinks=true]{hyperref}

\input{macros}

\iccvfinalcopy %

\def\iccvPaperID{5908} %

\ificcvfinal\fi

\begin{document}

\title{Learning Human-Human Interactions in Images from Weak Textual Supervision}

\author{Morris Alper and Hadar Averbuch-Elor\\
Tel Aviv University}

\maketitle
\ificcvfinal\thispagestyle{empty}\fi
\thispagestyle{plain}
\pagestyle{plain}

\begin{abstract}
   \input{00-abstract}

\end{abstract}

\section{Introduction}
\input{01-intro}

\section{Related Work}
\input{02-relatedwork}

\input{03-dataset}

\input{04-method}

\input{05-results}

\section{Conclusion}
\input{06-conclusion}

\input{07-ack}

{\small
\bibliographystyle{ieee_fullname}
\bibliography{egbib}
}

\clearpage
\appendix
{\LARGE\textbf{Appendix}}

\medskip
We refer readers to the interactive visualizations at \href{https://learning-interactions.github.io}{our project page} that show results for all presented models on the two test sets we examine (\datasetname{} and \emph{imSitu-HHI}). In this document, we describe additional experiments and results (Section \ref{sec:supp_results}) and provide additional details (Section \ref{sec:supp_details}).

\section{Additional Results and Comparisons}
\label{sec:supp_results}

\input{supp-1-03-coformer-imsitu}
\input{supp-1-04-extended-imsitu}
\input{supp-1-05-extractive-only}
\input{supp-1-06a-neural}
\input{supp-1-06b-ngram}
\input{supp-1-07-abl}
\input{supp-1-08-qualitative}

\section{Additional Details}
\label{sec:supp_details}

\input{supp-2-01-cc-news-captions}
\input{supp-2-02-parsing-interactions}

\input{supp-2-03-novel-interactions}
\input{supp-2-04-synthetic-generation}
\input{supp-2-05-pseudo-labels}

\input{supp-2-06-pseudo-labels-examples}
\input{supp-2-07-imsitu}
\input{supp-2-08-training-details}

\input{supp-2-09-baselines}
\input{supp-2-10-metrics}
\input{supp-2-11-coformer-eval}

\input{supp-3-img-attrib}

\end{document}

%% file: macros.tex
\usepackage{xcolor}
\usepackage{xspace}

\usepackage{multirow}
\usepackage{tabularx}

\usepackage{soul}

\newcommand{\cancel}[1]{{\textcolor{red}{\st{#1}}}}

\newcommand{\datasetname}{\emph{Waldo and Wenda}\xspace}

\newcommand{\imsituhhi}{\emph{imSitu-HHI}\xspace}

\long\def\ignorethis#1{}

\newcommand{\NAME}{\texttt{[NAME]}\xspace}
\newcommand{\SNAME}{\texttt{[*]}\xspace}

\newcommand{\whitetxt}[1]{{\color{white}#1}\normalfont}

\definecolor{darkpurple}{RGB}{102,0,102}
\definecolor{gold}{RGB}{255,128,0}
\definecolor{mediumaquamarine}{RGB}{102,205,170}
\definecolor{blueviolet}{RGB}{138,43,226}
\definecolor{cornflowerblue}{RGB}{100,149,237}
\definecolor{lightpink}{RGB}{255,182,193}
\definecolor{orchid}{RGB}{218,112,214}

\newbox\jsavebox
\newcommand{\jsubfig}[2]{%
	\sbox\jsavebox{#1}%
	\parbox[t]{\wd\jsavebox}{\centering\usebox\jsavebox\\#2}%
	}

%% file: 00-abstract.tex
Interactions between humans are diverse and context-dependent, but previous works have treated them as categorical, disregarding the heavy tail of possible interactions. We propose a new paradigm of learning human-human interactions as free text from a single still image, allowing for flexibility in modeling the unlimited space of situations and relationships between people. To overcome the absence of data labelled specifically for this task, we use knowledge distillation applied to synthetic caption data produced by a large language model without explicit supervision. We show that the pseudo-labels produced by this procedure can be used to train a captioning model to effectively understand human-human interactions in images, as measured by a variety of metrics that measure textual and semantic faithfulness and factual groundedness of our predictions. We further show that our approach outperforms SOTA image captioning and situation recognition models on this task. We will release\footnote{via our project page \url{https://learning-interactions.github.io}} our code and pseudo-labels along with \textbf{\datasetname{}}, a manually-curated test set for still image human-human interaction understanding.

%% file: 01-intro.tex
\emph{``No man is an island entire of itself."}
\qquad  \emph{-John Donne}\\

Humans are social beings. As such, interactions among people are ubiquitous and diverse, affected by various factors including social context and cultural norms. Reasoning about these interactions is crucial for gaining a holistic understanding of visual scenes depicting people. However, in spite of significant progress in analyzing isolated human actions \cite{kong2022human, yadav2021review, zhang2019comprehensive} and relationships between entities and objects \cite{kim2021hotr, zou2021end}, far less attention has been devoted towards an automatic understanding of human-human interactions (HHI). This is despite the importance of this task for applications such as interactive robotics, social behaviour understanding, and captioning systems for the visually impaired.

There are a number of factors that make the analysis of HHI difficult. The space of possible interactions between people is vast and requires understanding social context and physically non-local relationships, as illustrated in Figure \ref{fig:teaser}. In addition, images depicting HHI may have multiple interpretations, some of which may be simultaneously correct. For example, the image on the left might depict ``celebrating a wedding'' as well as ``dancing''. Contextual cues such as the cake in the background of the image provide additional information that hints at the depicted HHI.

\input{figures/teaser/teaser}

Prior works targeting HHI understanding focus on a small fixed number of interactions; representative works include \cite{tanisik2021multi, van2018hands, nour2014human, kong2014interactive}, all of whose models are trained to recognize no more than ten interaction classes. In this work, we are interested in modeling the heavy tail of possible HHI to better understand the rich variety of ways in which people interact. To this aim, we propose to model HHI understanding as free text generation; since HHI are not confined to a fixed set of categories or even to a syntactic class such as verbs, HHI as free text enables the expression of an infinite variety of possible interactions. Furthermore, in contrast to previous works that frequently rely on extra context such as video data~\cite{stergiou2019analyzing}, we use a \emph{single} image with no additional information (during inference), making our method more widely applicable.
We focus on what Stergiou and Poppe~\cite{stergiou2019analyzing} term \textit{dyadic} interactions---pairwise interactions between two people. Our goal is to identify the most salient dyadic interaction given an image of two or more people interacting.

One of the primary challenges to modeling HHI is a scarcity of labelled data for this particular task. There are only a handful of relatively small datasets specific to HHI, and larger video datasets for action recognition are lacking in coverage of interactions (see Table \ref{tab:datasets}). To better model the heavy tail of possible HHI, we leverage the abundance of high-quality images of people and associated textual captions available on the Internet. In particular, we use the Who's Waldo dataset~\cite{cui2021s} that contains 270K image-caption pairs from Wikimedia Commons depicting people captured in a broad range of situations. Unlike many other image captioning datasets, Who's Waldo focuses on human-centric situations which are described using real-world captioned Internet data, and thus is more relevant to HHI understanding. However, it is extremely challenging to learn HHI from raw Internet captions directly, due to significant noise introduced by clutter and irrelevant details. To overcome this, we infer interactions from the original captions by applying knowledge distillation to synthetic data generated by a large language model, without explicit supervision. This approach allows for creating accurate pseudo-labels that provide textual descriptions of the HHI depicted in the images. We will release these pseudo-labels along with a manually annotated test set containing 1K image-interaction pairs from diverse Internet images which we name \datasetname, a new benchmark for our paradigm of HHI understanding as free text on still images, capturing the heavy tail of human-human interactions.

We demonstrate the utility of these pseudo-labels for learning HHI from images by training captioning models and using them as targets for a language modelling objective. We provide qualitative and quantitative analysis on the \datasetname test set; in addition, we evaluate this method on a larger scale by applying it to verb prediction on the imSitu situation recognition dataset~\cite{yatskar2016}, which we filter to select for images relevant to HHI.

Because we predict HHI as free text rather than categorically as in previous works, we propose a set of evaluation metrics chosen to measure important aspects of predicted HHI quality, namely textual similarity, factual groundedness, and verb similarity. Our evaluation shows that our HHI pseudo-labels allow for generating meaningful HHI free text descriptions from images, as measured by these metrics. We also show that learning on these pseudo-labels captures HHI substantially more effectively than either using existing SOTA image captioning models as-is or than training on interactions extracted with naive syntactic parsing.
Explicitly stated, our key contributions are:
\begin{itemize}
    \item A new paradigm and benchmark for HHI understanding from images---\emph{i.e.}, predicting interactions as free text---allowing to better understand the vast variety of ways in which people interact.
    \item A method for isolating HHI from noisy Internet captions using knowledge distillation applied to a large language model, and a set of pseudo-labels generated by this method. 
    \item An evaluation framework with metrics that capture HHI understanding, and results demonstrating that training image captioning models on these pseudo-labels can allow for modeling the heavy tail of possible HHI across various situations and configurations more effectively than SOTA image captioning and situation recognition models.
\end{itemize}

%% file: figures/teaser/teaser.tex
\begin{figure}
    \centering
    \jsubfig{\includegraphics[height=2.75cm]{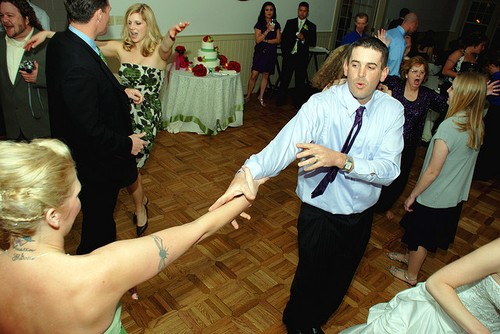}}{}
    \hfill
    \jsubfig{\includegraphics[height=2.75cm]{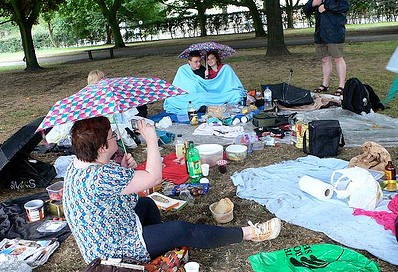}}{}
    \caption{How would you describe the interactions depicted in these images? There are unlimited possible interactions between people which cannot be easily described by a fixed set of categories or actions. Context plays a crucial role, as in the left image where the clothing and cake in the background help to interpret the depicted interaction. Moreover, interactions may involve participants at a physical distance as in the image on the right. To model the heavy tail of possible interactions, we propose to learn HHI as free text (see below\setcounter{footnote}{1}\protect\footnotemark{} for predictions using our method).}
    \label{fig:teaser}
\end{figure}

\footnotetext{A model fine-tuned on our pseudo-labels yields ``dancing'' and ``having a picnic''.} 

%% file: 02-relatedwork.tex
\noindent
\textbf{Human action recognition.}  Human actions span a range from simple to complex. These include simple actions (``running''), human-object interactions (``dribbling a ball''), human-human interactions (``shaking hands''), and group actions (``gathering''). Because of the dynamic nature of actions, a large portion of work on action recognition uses video data~\cite{tang2012learning,tran2015learning,carreira2017quo,zhu2020comprehensive,wang2021actionclip}. Other approaches use other modalities such as depth or skeleton data \cite{ye2013survey, presti20163d, yadav2021review}. Among video-based approaches, some use shallow approaches separating feature representation of action videos and classification of these features, while others use end-to-end trainable networks (see \cite{kong2022human,zhang2019comprehensive} for detailed surveys). Works on human-object interactions (HOI) may use separate modules such as human and object detectors and relation modules~\cite{chao2018learning,gkioxari2018detecting,gao2018ican}, pose and gaze estimation~\cite{li2019transferable,wan2019pose,xu2019interact}, or graph neural networks applied to scene graphs~\cite{qi2018learning,xu2019learning,zhou2019relation,liang2021visual}. One line of recent work on HOI uses end-to-end models, frequently with transformer architectures~\cite{tamura2021qpic,zou2021end,chen2021reformulating,kim2021hotr,chen2021qahoi}. In our work, we aim to predict the most salient interaction between the pictured individuals in an end-to-end manner from still image data alone.

HHIs are a subset of human actions which pose particular challenges to automatic recognition, due to non-locality, context dependency, and ambiguity. A number of works have explicitly tackled HHI recognition, as surveyed by Stergiou and Poppe \cite{stergiou2019analyzing}. As with general action recognition, these approaches most commonly use video data as input~\cite{nour2014human,gemeren2016spatio,wang2017recurrent,shu2019hierarchical,lee2022human}. However, a few works have tackled the more challenging task of HHI recognition in still images. Some of these use classical computer vision methods to estimate human locations and poses in photos for predicting HHI~\cite{yang2012recognizing,chakraborty20133d,Antol2014}. Xiong \emph{et al.}~\cite{xiong2015recognize} use a CNN architecture with human, face, and object detection features for event recognition. These works all treat HHI as categorical, predicting them from a small set of predefined interaction classes. In contrast, we use free text to describe HHIs allowing for more flexibility than categorical recognition.

\medskip \noindent
\textbf{HHI datasets.} 
Most existing datasets of HHI or with subsets representing HHI classes only include a small number of interaction categories. The majority consist of video data, either curated~\cite{ryoo2010ut,yun2012two,gemeren2016spatio,gu2018ava} or YouTube-based~\cite{smaira2020short,monfort2019moments,zhao2019hacs}.

There are few image datasets dedicated to human actions, of which HHI are a subset. Ronchi and Perona~\cite{ronchi2015describing} introduce the Visual Verbnet dataset consisting of images with dense verb annotations. %
Yatskar \emph{et al.}~\cite{yatskar2016} introduce the imSitu dataset for image situation recognition, involving recognizing the action portrayed in a still image (often with a human participant or participants) as well as predicting semantic roles for observed entities. In both cases the labels are selected from a fixed set of categories---single verbs in the case of imSitu, verbs or phrases containing verbs (\emph{e.g.} ``shake hands'') for Visual Verbnet. Other image datasets such as Visual Genome~\cite{krishna2017visual} contain labeled entities, objects and their relationships, but focus more on general objects rather people and their interactions.

See Table \ref{tab:datasets} for a comparison of the most related datasets with our proposed HHI dataset. Unlike prior datasets, ours represents HHI as free text and not as fixed categories.

\input{tables/datasets}

\medskip \noindent
\textbf{In-context learning with large language models (LLMs).}
The recent explosive growth in size and NLP benchmark performance of LLMs has led to their use as foundation models for use on downstream tasks~\cite{bommasani2021opportunities}. Models such as GPT-3 show an emergent \emph{in-context learning} property, whereby they may solve new tasks when prompted with only a few examples of a new task, or even just with a task description, without any parameter updates~\cite{brown2020language, dong2022survey}. The output of such models may then be used as supervised training data for conventional model fine-tuning. The idea of training on data generated using in-context learning to create a large training data set has been successfully applied to achieve state-of-the-art results on the SuperGLUE NLP benchmark by Wang \emph{et al.}~\cite{wang2021towards}. In our case, we use this data to perform \emph{sequence-level knowledge distillation} -- transferring the knowledge exhibited by such a large model into a smaller model by training on its output sequences~\cite{kim2016sequence,gou2021knowledge}.

The use of LLM-generated synthetic data for multimodal learning has been explored by Brooks \emph{et al.}~\cite{brooks2022instructpix2pix}, who use caption pairs generated by GPT-3 as auxiliary data for training a conditional diffusion model to perform image editing. Their method uses hundreds of manually labelled pairs of texts as training data; however, our pseudo-labelling method uses no explicit supervision, instead using syntactic parsing to generate automatic seeds for our synthetic data generation pipeline.

%% file: tables/datasets.tex
\begin{table}[t]
  \centering
  \setlength{\tabcolsep}{3.2pt}
  \def\arraystretch{0.95}
  \begin{tabularx}{0.999\columnwidth}{lcc}
    \toprule
    Dataset       & \#Seq      &  \#HHI Classes    \\
    \midrule
    \textbf{Curated videos} \\
    UT-Interaction~\cite{ryoo2010ut} & 60  & 6 \\
    TV Human Interaction~\cite{patron2010high} & 300  & 4 \\ %
    Hollywood2~\cite{marszalek2009actions} & 3669 &  4 \\
    ShakeFive2~\cite{gemeren2016spatio} & 153 &  5 \\
    SBU Kinect~\cite{yun2012two} & 300 &  8 \\
    AVA~\cite{gu2018ava} & $\sim$57.6k &  13 \\
    NTU RGB+D (120)~\cite{shahroudy2016ntu,liu2019ntu} & $\sim$114k &  26 \\
    \midrule
    \textbf{YouTube-based videos} \\
    Kinetics~\cite{kay2017kinetics,smaira2020short} & $\sim$500k &  11 \\
    Moments in Time~\cite{monfort2019moments} & $\sim$800k &  32 \\ %
    HACS~\cite{zhao2019hacs} & $\sim$50k &  23 \\
    \midrule
    \textbf{Still images} \\
    imSitu~\cite{yatskar2016} & ~126k &  $50^*$ \\
    Visual Verbnet~\cite{ronchi2015describing} & ~10k &  $52^*$ \\
    Who's Waldo~\cite{cui2021s} (\emph{w/ our labels}) & ~127k  & $\infty ^*$ (free text) \\
    \bottomrule
  \end{tabularx}
  \vspace{-10pt}
  {\begin{flushleft}
  \footnotesize $^*$The number of HHI classes for Visual Verbnet includes verbs in the \emph{communication}, \emph{contact} and \emph{social} categories, which sometimes mark solo actions or human-object interactions. The imSitu dataset contains a total of 504 verbs. We estimate the number of HHI interactions using an automatic methodology detailed in Section \ref{sec:results}. Our free text pseudo-labels are limited to the types of interactions available in Who's Waldo.
  \end{flushleft}}
  \vspace{-5pt}
  \caption{\textbf{Comparison of HHI datasets}. Prior datasets usually capture video data and target a small number of interaction classes. Several datasets focus on human actions, some of which include HHI. We denote the number of video/image samples with \#Seq, and the number of HHI classes with \#HHI Classes (values are taken from Stergiou and Poppe~\cite{stergiou2019analyzing} where relevant). In our work, we devise a technique for generating HHI pseudo-labels for Who's Waldo~\cite{cui2021s}, a dataset containing real-world image--caption pairs, allowing for modeling the heavy tail of HHI.
  }
\label{tab:datasets}
\end{table}

%% file: 03-dataset.tex
\input{figures/wenda/examples}

\section{LLM-Based HHI Inference from Captions}\label{sec:llm_based}

To model the heavy tail of possible HHI using free text, leverage weak supervision in the form of image captions. We turn to Who's Waldo~\cite{cui2021s}, a dataset containing image--caption pairs depicting human-centric scenes scraped from Wikimedia Commons (with names masked using their suggested \NAME token). As illustrated in Figure \ref{fig:examples}, the mentions of the depicted HHI are embedded in detailed textual captions, and do not directly correspond to syntactic structures such as verbs in the text. For instance, the first depicted caption is long and the only relevant detail is the phrase ``gets at [sic] high five''; the last depicted caption contains no verb (while the noun phrase ``Ski Tour'' hints at the relevant interaction). These captions are thus inadequate for training an HHI understanding model directly, as a captioning model fine-tuned on them mainly learns to attend to details that are irrelevant for our task (as shown in Section \ref{sec:ablations}). We therefore present a large language model (LLM)-based abstractive text summarization technique that produces clean interaction texts from the original Internet captions, without explicit supervision.

Our unsupervised pseudo-labelling approach operates in three stages, illustrated in Figure \ref{fig:pipeline}: (1) We extract syntactic parsing-based interactions from captions from the Who's Waldo dataset, as well as constructing new synthetic interaction texts. (2) We prompt an LLM using the interaction--caption pairs from Who's Waldo along with the new interactions. The output synthetic captions are filtered using a pretrained natural language inference (NLI) model and various textual heuristics, to select for those that correspond to the new interactions. (3) We train an abstractive summarization model on these synthetic caption--interaction pairs; this model learns to output HHI from noisy Internet captions. As seen in Figure \ref{fig:examples}, these interaction pseudo-labels accurately describe the HHI visible in their associated images. Below, we provide more details for each stage (Sections \ref{sec:distill1}--\ref{sec:distill3}). We then present \datasetname{}, our manually-curated HHI test set, in addition to statistics and an ethical discussion (Section \ref{sec:dataset}).

\subsection{Constructing interaction texts} \label{sec:distill1}
We first define a rule-based approach for extracting interactions via syntactic parsing. Specifically, we extract the first verb in the caption with a \NAME{} subject along with its direct objects and the heads of its prepositional arguments. This roughly corresponds to an interaction, although it may sound unnatural. This is also limited to captions containing verb phrases. We apply this procedure to captions from Who's Waldo to obtain corresponding parsing-based interactions.

We also construct new synthetic interactions by first applying this parsing procedure to scraped texts of news articles from the CC-News dataset~\cite{Hamborg2017} (from Common Crawl, containing text without image data), and then using the output interaction texts to prompt the large (1.3B-parameter) language model GPT-Neo~\cite{gpt-neo,gao2020pile}, which produces a set of diverse and more natural-sounding interactions.

\input{figures/pipeline/pipeline}

\subsection{Synthetic caption data generation}\label{sec:distill2}

Using the caption--interaction pairs from Who's Waldo and the new synthetic interactions as seeds, we generate synthetic caption--interaction pairs using in-context learning with GPT-Neo. This allows us to create a larger and more diverse set of caption--interaction pairs than by using caption--interaction pairs directly from Who's Waldo. These pairs serve as the teacher model outputs used for knowledge distillation in the following section.

At each step, the language model is shown a prompt beginning with multiple randomly-selected examples of caption--interaction pairs from Who's Waldo. This provides context for the model to understand the task at hand--associating interactions with captions that contain them. We use ten examples in each prompt to balance between the providing sufficient context with computational considerations. The prompt ends with a new desired interaction, and the language model proceeds to generate a caption corresponding to this interaction. We filter these results using a pretrained NLI model and various textual heuristics detailed further in the supplementary material, ensuring that the output caption logically is properly formatted and logically entails the corresponding interaction.

\subsection{Knowledge distillation for summarization}\label{sec:distill3}

Using the synthetic data generated in the previous stage, we fine-tune a smaller (220M-parameter) student T5 model, a sequence-to-sequence transformer network whose pretraining tasks include text summarization~\cite{raffel2020exploring}. We use the synthetic captions (with the task prefix ``summarize:'') as input and the synthetic interactions of the target text for fine-tuning. Empirically, we find that our fine-tuned student model is able to summarize captions and output valid interactions even when the caption does not contain a verb or has a syntactic structure that the syntactic parsing-based method could not process.

We apply this model to the captions in the Who's Waldo dataset to create pseudo-labels representing interactions as free text. See Figure \ref{fig:examples} for examples of such pseudo-labels.

\subsection{Our HHI dataset} \label{sec:dataset}
Using our learned abstractive summarization model, we may generate interaction pseudo-labels from Who's Waldo captions. Out of the ${\sim}270k$ samples in Who's Waldo, we use only those ${\sim}130k$ containing at least two human face detections, using the detections provided by Cui \emph{et al.}~\cite{cui2021s}. We filter out duplicate and near-duplicate images, those with high similarity to test set images, and samples with pseudo-labels that do not pass a few simple text-based filtering rules, including enforcing the format of \NAME{}, followed by a present continuous verb (``-ing''), and including another \NAME{} token. We are left with ${\sim}126k$ images with pseudo-labels in total, which we hereby refer to as pHHI.

\medskip 
\noindent \textbf{The \datasetname{} Benchmark.} %
We also create \datasetname{}\footnote{Wenda appears in the \emph{Where's Waldo?} book series as Waldo's girlfriend.}, an HHI test set containing 1K manually curated image--interaction text pairs. In order to test generalization to HHI understanding across a wide variety of natural images, we include data from three sources: (1) 300 images from Who's Waldo, (2) 300 images from COCO Captions~\cite{chen2015microsoft}, (3) 400 images from Conceptual Captions~\cite{sharma2018conceptual}. The images are selected from the validation and test splits of the relevant datasets. As the distribution of HHI in natural photographs is highly imbalanced---for instance, images in captioning datasets often display people standing side by side and posing for photographs---we curate this test set to represent a wide variety of interactions and to reflect performance on the long tail of uncommon HHI. Examples of images from \datasetname{} can be seen in Figures \ref{fig:teaser}, \ref{fig:examples}, and \ref{fig:results_wenda}.

\medskip \noindent \textbf{Dataset Statistics.}
Overall, our pHHI training dataset contains 126,696 pairs of images and pseudo-labels. These labels contain 1,263 unique verbs and 16,136 unique interactions. The majority of the images ($59.3\%$) only contain two detected people, with less than $5\%$ of the images containing more than six detected people. The \datasetname{} test set contains 1,000 images along with their manually written ground truth HHI labels. These include 238 unique verbs and 575 unique interaction labels.

\medskip \noindent \textbf{Ethical considerations.} Our dataset inherits a diverse representation of people (ages, ethnicities, geographic etc.) from the Who's Waldo dataset~\cite{cui2021s}. Furthermore, we use their provided name masking to mitigate biases (\emph{e.g.}, gender biases). We verify that all manually-curated test samples are neutral in nature and do not contain lurid or negative material. We perform similar verification on external test data, as described in Section \ref{sec:results}, to avoid exposure to harmful or offensive behaviors. Furthermore, our pseudo-labels and test set will only be made available for academic purposes.

%% file: figures/wenda/examples.tex
\begin{figure*}
\jsubfig{\includegraphics[height=2.7cm]{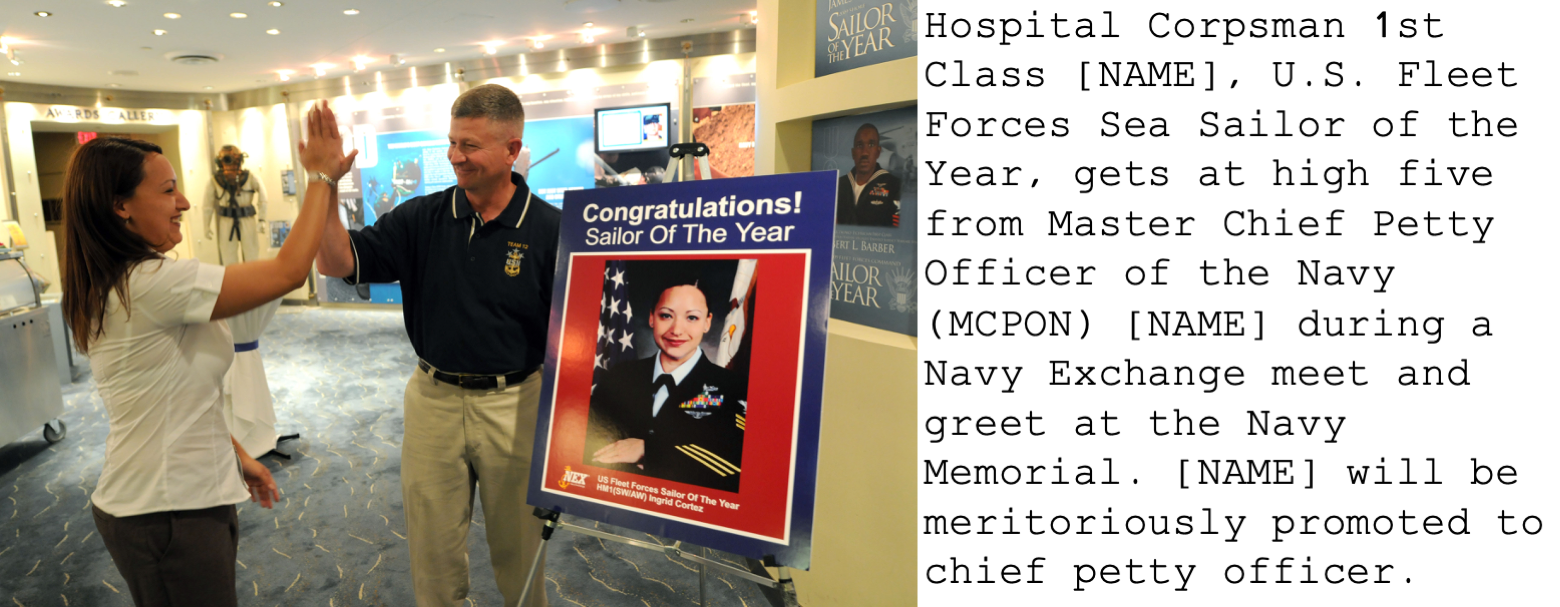}}{\vspace{-10pt}\begin{flushleft}\small \emph{Parsing-based:}{\SNAME{} getting five from \SNAME{}} \\ \emph{Final:}\hspace{33px}{\SNAME{} getting a high five from \SNAME{}} \end{flushleft}} 
\hfill
\jsubfig{\includegraphics[height=2.7cm]{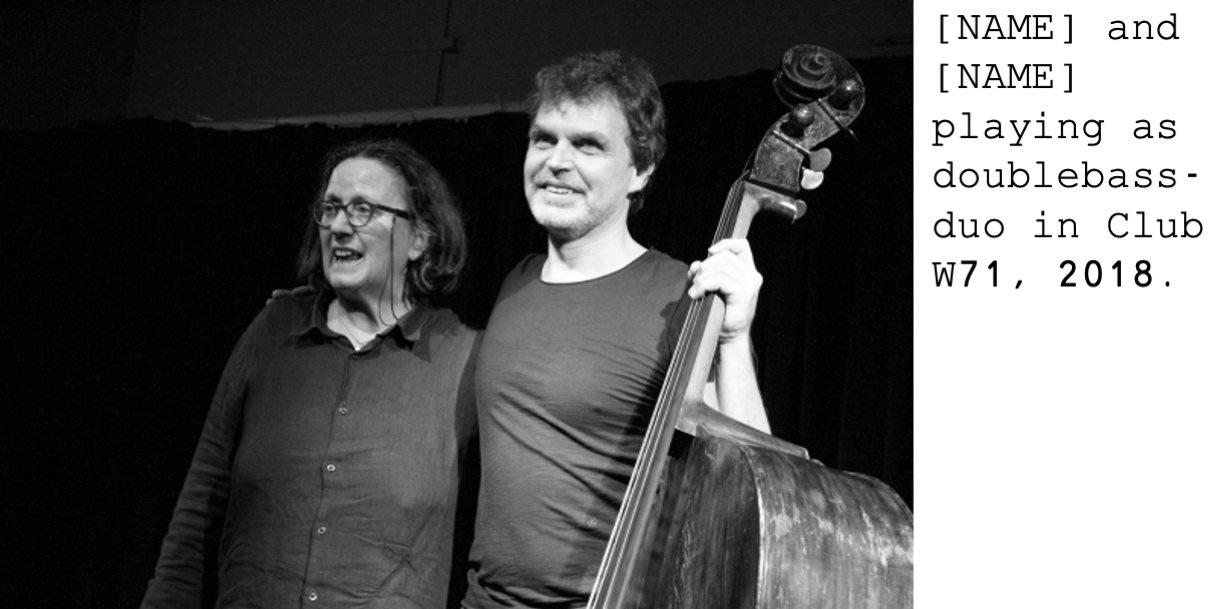}}{\vspace{-10pt}\begin{flushleft} \small \SNAME{} and \SNAME{} playing \\
{\SNAME{} playing doublebass with \SNAME{}}
\end{flushleft}} %
\hfill
\jsubfig{\includegraphics[height=2.7cm]{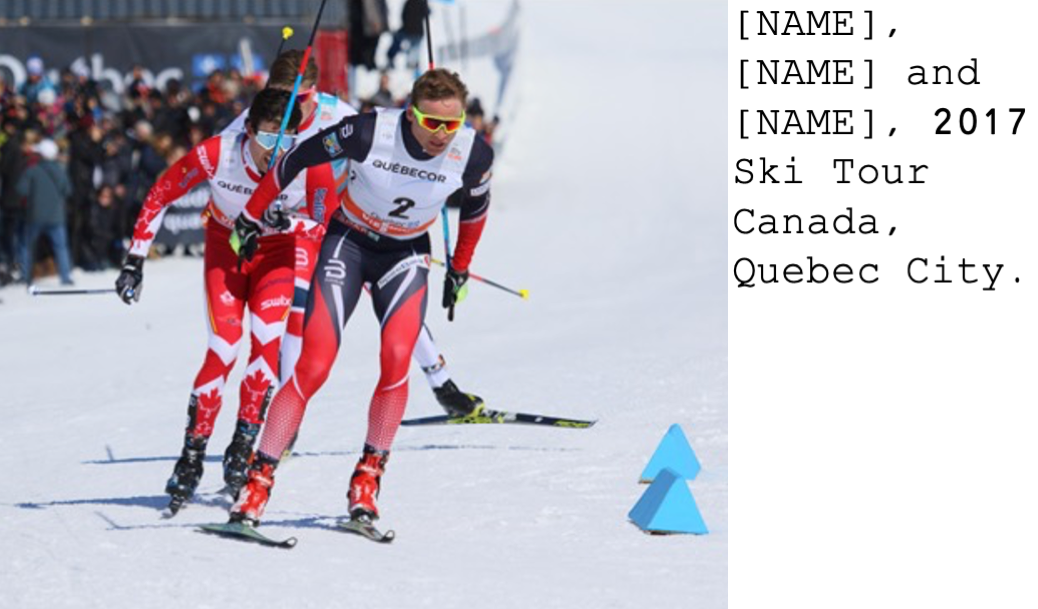}}{\vspace{-10pt}\begin{flushleft} \small $\emptyset$\whitetxt{\NAME{}} \\ {\SNAME{} skiing with \SNAME{}}\end{flushleft}} 
\vspace{-6pt}
  \caption{ \textbf{{HHI} distilled from {raw Internet captions} alongside their corresponding images.} On top we show several images and captions from the Who's Waldo dataset~\cite{cui2021s}, with \SNAME{} denoting masked named person entities. Our syntactic parsing approach allows for extracting an initial partial set of interactions (first row). We refine and enlarge this initial set using our abstractive summarization model which yields our final HHI pseudo-labels (second row).
  While the original captions possibly contain many additional details or no verb-based interaction at all (for example, see the rightmost image), our abstractive HHI pseudo-labels succeed in describing HHI visible in the associated images.
  }\label{fig:examples}
\end{figure*}

%% file: figures/pipeline/pipeline.tex
\begin{figure*}

\jsubfig{\includegraphics[height=2.4cm]{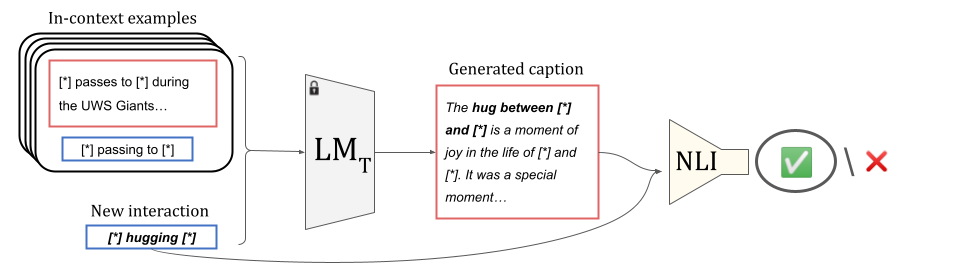}}
{\vspace{-10pt}\begin{flushleft}\small \centering Synthetic caption generation with teacher LM and NLI filtering\end{flushleft}} 
\hfill
\jsubfig{\includegraphics[height=2.4cm]{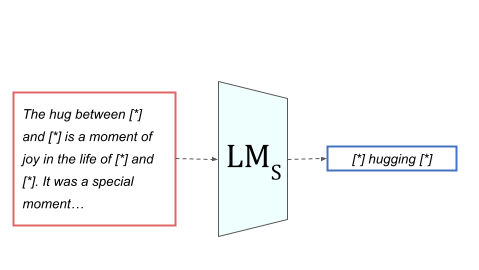}}
{\vspace{-10pt}\begin{flushleft}\small  \centering Fine-tuning student LM\end{flushleft}} 
\hfill
\jsubfig{\includegraphics[height=2.4cm]{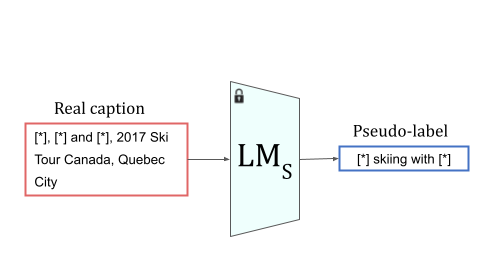}}
{\vspace{-10pt}\begin{flushleft}\small \centering Pseudo-label inference\end{flushleft}} 
\hfill
    \caption{\textbf{LLM-Based HHI Extraction from Captions.} We generate synthetic interaction-caption pairs via in-context learning (left), use them to fine-tune a summarization model (center), and then use this model to producing HHI pseudo-labels for captions in Who's Waldo (right), as detailed in Section \ref{sec:llm_based}. Captions are shown in red boxes, interaction texts in blue, and synthetic texts in italic letters. $\text{LM}_{\text{T}}$ and $\text{LM}_{\text{S}}$ indicate teacher and student language models respectively.}
    \label{fig:pipeline}
\end{figure*}

%% file: 04-method.tex
\input{figures/results/wenda/wenda}

\section{Learning HHI from Still Images}
\label{sec:method}

In the previous section, we demonstrated how we can obtain free text HHI pseudo-labels from the Internet captions of the Who's Waldo dataset~\cite{cui2021s}. We proceed to show how we use these to supervise learning HHI from still images via the paradigm of image captioning.

\subsection{Models considered}

After obtaining a set of images and pseudo-labels, we consider the task of HHI in the framework of image captioning. Given (image, pseudo-label) pair $(I, L)$, we train an encoder-decoder network $M$ to maximize the predicted conditional likelihood of $L$ using a cross-entropy loss. During inference, we use autoregressive beam search decoding to generate text token by token, given $I$ as input.

In order to evaluate the utility of transfer learning from general image captioning to our HHI understanding setting, we evaluate two choices for the model $M$:

\medskip \noindent
\textbf{(1) Vanilla encoder-decoder (EncDec).} In this setting, we fine-tune a simple encoder-decoder model. We use the image encoder of pretrained CLIP-ViT~\cite{radford2021learning}, with its pooled embedding output followed by a single linear projection layer to match the hidden dimension of the decoder. For the decoder we use pretrained GPT-2~\cite{radford2019language} with a causal language modelling head and cross-attention over the encoder output. Consistent with previous works on fine-tuning vision-and-language models \cite{luo2022frustratingly,mokady2021clipcap,zhai2022lit}, we freeze the weights of the image encoder as we fine-tune it on our pHHI data. 
By considering this model that was not previously trained on image captioning, we aim to evaluate the extent to which our pHHI aid in learning to understand the semantics of HHI in images (rather than simply cueing a captioning model to the correct surface form of HHI labels).

\medskip \noindent
\textbf{(2) Fine-tuned captioner.} The second approach we consider is to apply transfer learning to a SOTA captioning model by fine-tuning it on our pHHI data. Because the Conceptual Captions (CC) dataset is more people-centric than COCO and thus closer to our use case, we pick CLIPCap~\cite{mokady2021clipcap} pretrained on CC as the base model for fine-tuning.  Consistent with CLIPCap's training method, we freeze its image encoder and fine-tune the model on our pHHI data.

\subsection{Training and Decoding}

For all models, we use cross-entropy loss and consistent hyperparameter settings. For each model, we decode using beam search with 32 beams. We report metric values for the top $1$, $5$, and $8$ beams.

%% file: figures/results/wenda/wenda.tex
\begin{figure*}
  \jsubfig{\includegraphics[height=2.88cm]{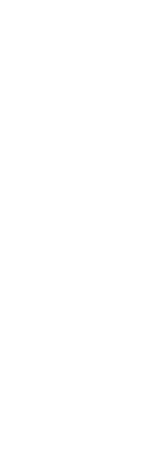}}{\begin{flushleft}\vspace{-21pt} \small{ CLIPCap \newline (CC)  \newline \vspace{6pt}  \newline   EncDec \newline (pHHI) \newline \vspace{-3pt} \newline GT}
  \end{flushleft}} 
  \hfill
  \hfill
  \jsubfig{\includegraphics[height=2.88cm]{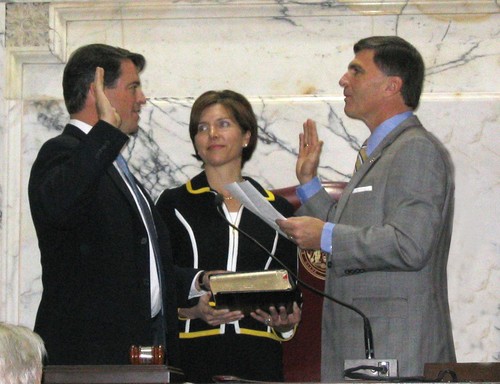}}{\vspace{-10pt}\begin{flushleft} \small{  person, left, and person, right, receive a standing ovation for their service.\\ \vspace{5pt}     \SNAME{} administering the oath to \SNAME{} \\ \vspace{8pt}     \SNAME{} swearing in \SNAME{}} \end{flushleft}}
          \hfill
  \jsubfig{\includegraphics[height=2.88cm]{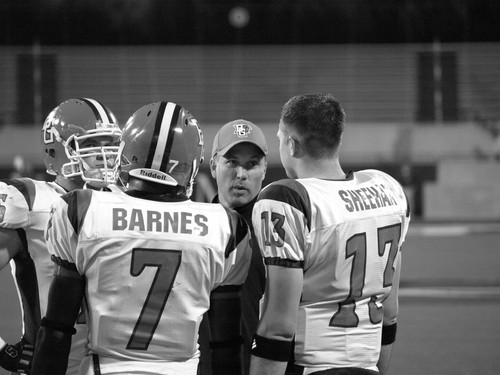}}{\vspace{-10pt}\begin{flushleft} \small{ friday night rivals was for high school vs game! \\ \vspace{16pt} \SNAME{} coaching \SNAME{}  \\ \vspace{20pt}  \SNAME{} huddling with \SNAME{} } \end{flushleft}}
          \hfill
\jsubfig{\includegraphics[height=2.88cm]{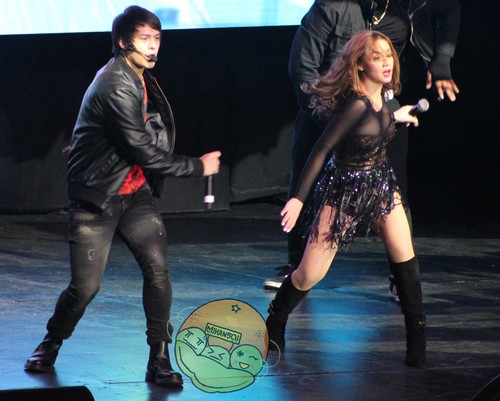}}{\vspace{-10pt}\begin{flushleft} \small{ wallpaper with a concert and a well dressed person entitled pop artist.  \vspace{5pt}  \newline  \SNAME{} performing with \SNAME{}  \\ \vspace{20pt}     \SNAME{} dancing with \SNAME{} } \end{flushleft}}
          \hfill
  \jsubfig{\includegraphics[height=2.88cm]{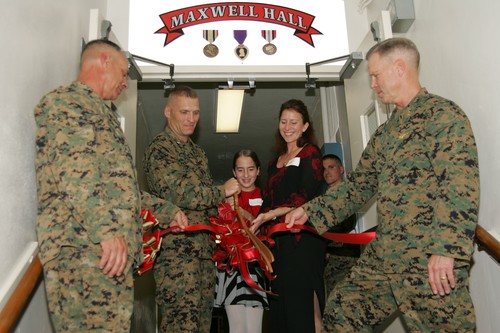}}{\vspace{-10pt}\begin{flushleft} \small{ person, left, shakes hands with person, daughter of person, during a ribbon cutting \vspace{5pt}  ceremony.   \SNAME{} cutting the ribbon with \SNAME{}  \\ \vspace{20pt}     \SNAME{} cutting a ribbon with \SNAME{} } \end{flushleft}}
\vspace{-3pt}
  \caption{\textbf{Results on the \datasetname{} test set.} We compare results obtained by a baseline, our vanilla encoder-decoder technique (trained on our pHHI data), and the ground truth labels in \datasetname{}, with \SNAME{} denoting \NAME{} tokens that represent person entities.  As illustrated above, our method generates text describing the HHI depicted in the image, without attending to other irrelevant details. In comparison, the SOTA captioning model CLIPCap used as-is may not output an interaction at all (middle two images). We also observe that our model predicts HHI that may require both a verb and other arguments to adequately understand (leftmost and rightmost images).
  }\label{fig:results_wenda}
\end{figure*}

%% file: 05-results.tex
\section{Evaluation}
\label{sec:results}

\input{05-01-test-datasets}

\input{05-02-baselines-ablations}

\input{tables/wenda_results}
\input{tables/wenda_source_results}

\input{05-03-metrics}

\input{tables/imsitu_results}

\input{05-04-discussion}

%% file: 05-01-test-datasets.tex
\subsection{Test Datasets}

We evaluate our models on the following datasets:

\medskip \noindent
\textbf{\datasetname{}.} As detailed in Section \ref{sec:dataset}, this consists of 1,000 images with manually-written ground truth labels. Examples of ground truth labels along with model predictions can be seen in Figure \ref{fig:results_wenda}. We report metric values averaged over the three data sources (Who's Waldo, Conceptual Captions, COCO) of \datasetname{} in Table \ref{tab:wenda_results}. We also show a breakdown of data source in Table \ref{tab:wenda_source_results}.

\medskip \noindent
\textbf{\imsituhhi.} We use an 8,021-sample subset of the imSitu~\cite{yatskar2016} situation recognition benchmark, which we refer to as \imsituhhi, to perform a large-scale evaluation of our models. Although imSitu does not contain free text HHI labels, it does contain categorical verb labels which can be used for comparison. Additionally, as the majority of images in imSitu do not depict HHI, we first filter for relevant samples as follows: We use person detections from YoloV5~\cite{glenn_jocher_2021_5563715} to select for images containing at least two humans. We further filter to select only samples with semantic frames containing at least two human participants. Finally, due to the noisy nature of this filtering, we only use verbs supported by at least $100$ images in this filtered subset, as these verbs are most likely to describe HHI. We use these verb labels as the ground truth and evaluate predictions with the verb similarity metric as described below.

%% file: 05-02-baselines-ablations.tex
\subsection{Baseline comparisons}

We compare our approach to two types of SOTA model of that do not use our pHHI data as baselines:

\textbf{(1) Pretrained captioner.} The first baseline approach that we test is the use of a SOTA model that has already been pretrained for image captioning. We test the recent captioning models ExpansionNet v2 (ENv2)~\cite{hu2022expansionnet} and CLIPCap~\cite{mokady2021clipcap}. We use these captioners as-is and evaluate our metrics on their outputs with beam search decoding. CLIPCap is avilable with pretrained weights for both COCO~\cite{chen2015microsoft} and Conceptual Captions (CC)~\cite{sharma2018conceptual}, and thus we test both models. ENv2 only uses COCO weights.

\textbf{(2) Pretrained situation recognition model.} We provide a comparison to the results of the CoFormer model~\cite{cho2022collaborative} for grounded situation recognition with pretrained weights. Unlike weakly-supervised models trained on our pseudo-labels, which were generated from natural captions, CoFormer is supervised by training on the manually-labelled SWiG dataset~\cite{pratt2020grounded}, an extension of imSitu which includes grounding information for arguments that are visible in the accompanying images, and the model predicts the relevant verb, arguments, and grounding information given an image.
We evaluate CoFormer by using its predicted verb, discarding semantic frame and grounding predictions since these semantic arguments do not directly map to the text of a human-human interaction string. %
See the supplementary material for details on how we insert its verb predictions into text prompts for metric calculations.

\subsection{Ablations} \label{sec:ablations}

In order to ablate the effect of our pseudo-labelling, we also report results of a captioning model fine-tuned on the entire text of the captions provided in Who's Waldo (listed in \ref{tab:wenda_results} and \ref{tab:imsitu_results} under training data as ``WW''). In the supplementary material we also provide a detailed comparison with results when training directly on the syntactic parsing-based seeds described in Section \ref{sec:distill1}.

%% file: tables/wenda_results.tex
\begin{table}[t]
  \centering
  \setlength{\tabcolsep}{4pt}
  \def\arraystretch{0.95}
  \begin{tabularx}{0.99\columnwidth}{llcccc}
    \toprule
    Method & Train Data & BL$\,\uparrow$ & $p_e \uparrow$ & $p_c \downarrow$ & sim$\,\uparrow$   \\
    \midrule
    \textbf{Results@1} \\
    CoFormer & SWiG & $\;\,$0.41$^*$ &	$\;\,$0.33$^*$ &	$\;\,$0.28$^*$ & 0.35 \\
    ENv2 & COCO & 0.27&	0.25&	\underline{0.33} & 0.41 \\
    CLIPCap &COCO  & 0.28&	\underline{0.34} &	0.37 & \underline{0.42}  \\
    CLIPCap &CC  & 0.27	&0.18&	0.38 & 0.35 \\
    CLIPCap &CC+WW & 0.26&	0.16&	0.40 & 0.17 \\
    EncDec & pHHI & \underline{0.38} & 0.30 & 0.37 & 0.41 \\
    CLIPCap &CC+pHHI & \textbf{0.42} &	\textbf{0.41} &	\textbf{0.32} & \textbf{0.46} \\
    \midrule
    \textbf{Results@5} \\
    ENv2 & COCO & 0.31 &	0.39 &	0.19 & 0.46 \\
    CLIPCap &COCO  & 0.31 &	0.47 &	0.24 & 0.46 \\
    CLIPCap & CC  & 0.33 &	0.32 &	0.20 & 0.47 \\
    CLIPCap &CC+WW & 0.33 &	0.29 &	0.24 & 0.27 \\
    EncDec & pHHI & \underline{0.51} & \underline{0.61} & \underline{0.09} & \underline{0.59} \\
    CLIPCap &CC+pHHI  & \textbf{0.57} &	\textbf{0.71} &	\textbf{0.07} & \textbf{0.65} \\
    \midrule
    \textbf{Results@8} \\
    ENv2 &COCO & 0.32 &	0.43 &	0.17 & 0.48 \\
    CLIPCap & COCO & 0.32 &	0.50 &	0.21 & 0.46 \\
    CLIPCap  & CC & 0.35 &	0.36 &	0.16 & 0.49 \\
    CLIPCap &CC+WW & 0.35 &	0.33 &	0.21 & 0.31 \\
    EncDec &pHHI & \underline{0.54}  & \underline{0.65} & \underline{0.19} & \underline{0.65} \\
    CLIPCap  & CC+pHHI &  \textbf{0.59} &	\textbf{0.76} &	\textbf{0.04} & \textbf{0.69} \\
    \bottomrule
  \end{tabularx}
  \vspace{-2pt}
  {\begin{flushleft}
  \footnotesize $^*$Evaluated by using the best of two prompt templates for each item, as described in the supplementary material.
  \end{flushleft}}
  \vspace{-4pt}
  \caption{Results on \textbf{\datasetname{}}. The listed metrics are BLEURT (BL) and NLI scores ($p_e, p_c$) and verb embedding similarity (sim). CC+WW/pHHI indicates models that were initialized with pretrained CC weights and subsequently fine-tuned on Who's Waldo captions or on pHHI respectively. Best results are in bold, and second best are underlined. Results are aggregated across the three data sources of \datasetname{}. For models using beam search, we report results for top 1, 5, and 8 beams.
  }
  \vspace{-8pt}
\label{tab:wenda_results}
\end{table}

%% file: tables/wenda_source_results.tex
\begin{table*}[t]
  \centering
  \setlength{\tabcolsep}{3.2pt}
  \def\arraystretch{0.95}
  \begin{tabularx}{0.96\textwidth}{llccccccccccccc}
    \toprule
    & &
    \multicolumn{4}{c}{\textbf{WW}} &
    \multicolumn{4}{c}{\textbf{CC}} &
    \multicolumn{4}{c}{\textbf{COCO}}
    \\
    \cmidrule(lr){3-6}
    \cmidrule(lr){7-10}
    \cmidrule(lr){11-14}
    Method & Train Data & BL$\,\uparrow$ & $p_e \uparrow$ & $p_c \downarrow$ & sim$\,\uparrow$ & BL$\,\uparrow$ & $p_e \uparrow$ & $p_c \downarrow$ & sim$\,\uparrow$ & BL$\,\uparrow$ & $p_e \uparrow$ & $p_c \downarrow$ & sim$\,\uparrow$  \\
    \midrule
    \textbf{Results@1} \\
    CoFormer & SWiG & $\;\,$0.40$^*$ & $\;\,$0.29$^*$ & $\;\,$0.35$^*$ & 0.34  & $\;\,$0.45$^*$ & $\;\,$0.40$^*$ & $\;\,$0.43$^*$ & 0.38 & $\;\,$0.37$^*$ & $\;\,$0.30$^*$ & $\;\,$0.33$^*$ & 0.33 \\
    Env2 & COCO & 0.24 & 0.20 & \textbf{0.28} & 0.38  & 0.26 & 0.20 & 0.45 & 0.41  & 0.31 & 0.36 & \textbf{0.26} & \underline{0.45}  \\
    CLIPCap & COCO & 0.27 & 0.37 & 0.35 & 0.39 & 0.26 & 0.26 & 0.43 & 0.43 & 0.31 & \underline{0.38} & 0.33 & 0.44  \\
    CLIPCap & CC & 0.30 & 0.24 & 0.42 & 0.36  & 0.25 & 0.18 & 0.38 & 0.40 & 0.26 & 0.11 & 0.34 & 0.30 \\
    CLIPCap & CC+WW & 0.28 & 0.12 & 0.59 & 0.27 & 0.26 & 0.21 & \underline{0.34} & 0.15 & 0.23 & 0.15 & \underline{0.28} & 0.10 \\
    EncDec & pHHI & \underline{0.41} & \textbf{0.38} & \underline{0.30} & \underline{0.42}  & \underline{0.38} & \underline{0.30} & 0.44 & \underline{0.42} & \underline{0.34} & 0.22 & 0.36 & 0.38 \\
    CLIPCap & CC+pHHI & \textbf{0.42} & \textbf{0.38} & 0.33 & \textbf{0.45}  & \textbf{0.44} & \textbf{0.44} & \textbf{0.33} & \textbf{0.47} & \textbf{0.40} & \textbf{0.40} & 0.30 & \textbf{0.47} \\
    \midrule
\textbf{Results@5} \\
Env2 & COCO & 0.28 & 0.30 & 0.17 & 0.42 & 0.30 & 0.32 & 0.28 & 0.46 & 0.35 & 0.55 & 0.12 & 0.51 \\
CLIPCap & COCO & 0.31 & 0.50 & 0.21 & 0.42  & 0.29 & 0.38 & 0.28 & 0.46 & 0.34 & 0.52 & 0.23 & 0.49 \\
CLIPCap & CC & 0.36 & 0.41 & 0.23 & 0.46 & 0.30 & 0.31 & 0.23 & 0.50 & 0.32 & 0.23 & 0.14 & 0.43 \\
CLIPCap & CC+WW & 0.33 & 0.20 & 0.50 & 0.33 & 0.34 & 0.37 & 0.13 & 0.28 & 0.33 & 0.31 & 0.09 & 0.20 \\
EncDec & pHHI & \underline{0.55} & \underline{0.61} & \underline{0.11} & \underline{0.63} & \underline{0.51} & \underline{0.65} & \underline{0.10} & \underline{0.59} & \underline{0.46} & \underline{0.56} & \underline{0.07} & \underline{0.56} \\
CLIPCap & CC+pHHI & \textbf{0.57} & \textbf{0.64} & \textbf{0.10} & \textbf{0.64} & \textbf{0.60} & \textbf{0.75} & \textbf{0.06} & \textbf{0.68} & \textbf{0.53} & \textbf{0.74} & \textbf{0.05} & \textbf{0.63} \\
\midrule
\textbf{Results@8} \\
Env2 & COCO & 0.29 & 0.34 & 0.15 & 0.42 & 0.31 & 0.35 & 0.25 & 0.48 & 0.36 & 0.59 & 0.10 & 0.53 \\
Env2 & COCO & 0.32 & 0.53 & 0.18 & 0.43 & 0.30 & 0.41 & 0.25 & 0.47 & 0.35 & 0.55 & 0.21 & 0.50 \\
CLIPCap & CC & 0.38 & 0.47 & 0.17 & 0.48 & 0.32 & 0.35 & 0.19 & 0.53 & 0.34 & 0.26 & 0.11 & 0.45 \\
CLIPCap & CC+WW & 0.34 & 0.22 & 0.48 & 0.35 & 0.36 & 0.42 & 0.10 & 0.33 & 0.35 & 0.34 & 0.06 & 0.25 \\
EncDec & pHHI & \textbf{0.60} & \underline{0.69} & \textbf{0.06} & \textbf{0.69} & \underline{0.55} & \underline{0.72} & \underline{0.05} & \underline{0.64} & \underline{0.50} & \underline{0.66} & \underline{0.04} & \underline{0.61} \\
CLIPCap & CC+pHHI & \textbf{0.60} & \textbf{0.70} & \underline{0.07} & \underline{0.68} & \textbf{0.63} & \textbf{0.81} & \textbf{0.03} &  \textbf{0.72} & \textbf{0.55} & \textbf{0.78} & \textbf{0.03} & \textbf{0.67} \\
    \bottomrule
  \end{tabularx}
  \vspace{-2pt}
  {\begin{flushleft}
  \footnotesize $^*$Evaluated by using the best of two prompt templates for each item, as described in the supplementary material.
  \end{flushleft}}
  \vspace{-4pt}
  \caption{Results on \textbf{\datasetname{} split by data source} -- Who's Waldo (WW), Conceptual Captions (CC), and COCO Captions. For models using beam search, we report results for top 1, 5, and 8 beams.}
  \vspace{-8pt}
\label{tab:wenda_source_results}
\end{table*}

%% file: 05-03-metrics.tex
\subsection{Metrics}

A number of metrics have been proposed for natural language generation tasks, measuring various aspects of text quality~\cite{he2022blind, gehrmann2023repairing}. As no prior works (to the best of our knowledge) predict HHI as free text, we propose a set of metrics that evaluate various relevant aspects of generated text:

\medskip \noindent
\textbf{Textual similarity.} We use the BLEURT~\cite{sellam2020bleurt} metric to measure similarity to the ground truth interaction. This is a learned metric for text generation which measures similarity between the text output by a model and the reference text. Because our test set is relatively small and the reference texts are short, this better reflects textual similarity than ngram-based metrics such as BLEU~\cite{papineni2002bleu} which have high variance and must be averaged over large datasets, as is shown in detail in the supplementary material.

\medskip \noindent
\textbf{Factual groundedness.} A key property of generated text is whether it is \emph{consistent} or \emph{contradictory} with respect to the ground truth (such as a source document in the case of summarization, or a reference caption in the case of image captioning)~\cite{kryscinski2019evaluating}. This may be quantified by using the scores output by a natural language inference (NLI) model, in order to measure the degree of factual groundedness or hallucination in generated text \cite{maynez2020faithfulness,laban2022summac}. For example, given an image with ground truth label \emph{\NAME sitting next to \NAME}, the prediction \emph{\NAME standing with \NAME} logically contradicts the reference label and thus is a factual hallucination. To measure this, we use scores ($p_e, p_c$) from a pretrained NLI model to estimate the factual groundedness of the predicted text, where $p_e$ is the probability of entailment and $p_c$ is the probability of contradiction. We treat the image caption from \datasetname{} as the premise and the model's prediction as the hypothesis for NLI inference. For test items sourced from COCO Captions, in which images correspond to multiple reference captions, we use the first reference as the premise for this calculation.

\medskip \noindent
\textbf{Verb similarity.} We calculate the average cosine similarity of the predicted and ground truth verbs in GloVe~\cite{pennington2014glove} embedding space. The motivation for this metric is that a prediction may be valid or nearly valid even if it is not identical to the ground truth label as long as the semantic distance between the verbs is small (e.g. ``hugging'' vs. ``embracing''). To evaluate this on free text predictions, we either select the first non-\NAME{} word in the output (for models trained on pHHI) or extract its first verb using a syntactic parsing model. If syntactic parsing does not yield a verb, the zero vector is used as the given embedding.

%% file: tables/imsitu_results.tex
\begin{table*}[t]
  \centering
  \setlength{\tabcolsep}{3.5pt}
  \def\arraystretch{0.95}
  \begin{tabularx}{1.0\textwidth}{llccccccccccccccccc}
     & &
    {\small \rotatebox[origin=l]{90}{\textbf{Average sim.}}} &
    &
    {\small\rotatebox[origin=l]{90}{socializing}} & {\small\rotatebox[origin=l]{90}{distributing} }& {\small\rotatebox[origin=l]{90}{teaching}} & {\small\rotatebox[origin=l]{90}{communicating}} & {\small\rotatebox[origin=l]{90}{interviewing}} & {\small\rotatebox[origin=l]{90}{lecturing}} &
    {\small\rotatebox[origin=l]{90}{training}} &
    {\small\rotatebox[origin=l]{90}{providing}} &
    {\small\rotatebox[origin=l]{90}{instructing}} &
    {\small\rotatebox[origin=l]{90}{giving}} &
    {\small\rotatebox[origin=l]{90}{pushing}} &
    {\small\rotatebox[origin=l]{90}{helping}} &
    {\small\rotatebox[origin=l]{90}{asking}} &
    {\small\rotatebox[origin=l]{90}{coaching}} &
    {\small\rotatebox[origin=l]{90}{talking}} \\
    \midrule
    \textbf{Results@1} \\
    ENv2 & COCO & 0.22 &  & \underline{0.19} & 0.07 & 0.22 & 0.21 & 0.28 & 0.19 & 0.25 & 0.26 & 0.11 & 0.45 & \textbf{0.38} & 0.29 & 0.44 & 0.28 & 0.28 \\
    CLIPCap & COCO & 0.23 &  & 0.18 & 0.07 & 0.26 & 0.24 & \underline{0.29} & 0.17 & 0.26 & 0.24 & 0.10 & 0.44 & 0.36 & 0.25 & \textbf{0.46} & 0.30 & 0.63 \\
    CLIPCap& CC  & 0.27 &  & 0.16 & \textbf{0.25} & \underline{0.37} & 0.26 & 0.21 & 0.23 & 0.28 & 0.35 & 0.16 & 0.46 & 0.31 & \underline{0.37} & 0.38 & 0.26 & 0.53 \\
    CLIPCap & CC+WW & 0.09 &  & 0.02 & 0.08 & 0.12 & 0.05 & \textbf{0.30} & 0.08 & 0.10 & 0.09 & 0.06 & 0.25 & 0.07 & 0.09 & 0.11 & 0.05 & 0.08 \\
    EncDec & pHHI& \underline{0.28} &  & \underline{0.19} & 0.21 & 0.34 & \underline{0.27} & 0.23 & \underline{0.24} & \underline{0.35} & \underline{0.38} & \underline{0.17} & \underline{0.60} & 0.36 & 0.34 & \textbf{0.46} & \textbf{0.76} & \underline{0.64} \\
    CLIPCap & CC+pHHI & \textbf{0.32} &  & \textbf{0.21} & \textbf{0.25} & \textbf{0.56} & \textbf{0.33} & 0.27 & \textbf{0.30} & \textbf{0.43} & \textbf{0.44} & \textbf{0.19} & \textbf{0.66} & \textbf{0.38} & \textbf{0.44} & \textbf{0.46} & \underline{0.65} & \textbf{0.70} \\
    \midrule
    \textbf{Results@5} \\
    ENv2 & COCO & 0.26 &  & 0.21 & 0.10 & 0.26 & 0.23 & 0.30 & 0.21 & 0.27 & 0.31 & 0.13 & 0.49 & 0.45 & 0.35 & 0.49 & 0.33 & 0.31 \\
    CLIPCap & COCO  & 0.25 &  & 0.19 & 0.09 & 0.29 & 0.26 & 0.31 & 0.20 & 0.27 & 0.26 & 0.12 & 0.47 & 0.41 & 0.29 & 0.48 & 0.32 & 0.66 \\
    CLIPCap & CC  & 0.35 &  & 0.21 & 0.30 & 0.48 & 0.33 & 0.28 & 0.31 & 0.37 & 0.43 & 0.21 & 0.56 & 0.42 & 0.48 & 0.47 & 0.31 & 0.64 \\
    CLIPCap & CC+WW & 0.18  &  & 0.05 & 0.15 & 0.22 & 0.14 & \textbf{0.60} & 0.16 & 0.17 & 0.22 & 0.11 & 0.46 & 0.15 & 0.21 & 0.22 & 0.15 & 0.24 \\
    EncDec &pHHI & \underline{0.40} &  & \textbf{0.30} & \textbf{0.35} & \underline{0.49} & \underline{0.41} & 0.39 & \underline{0.33} & \underline{0.51} & \underline{0.51} & \underline{0.23} & \underline{0.79} & \underline{0.47} & \underline{0.49} & \underline{0.57} & \textbf{0.92} & \underline{0.86} \\
    CLIPCap & CC+pHHI & \textbf{0.44} &  & \underline{0.29} & \textbf{0.35} & \textbf{0.85} & \textbf{0.44} & \underline{0.41} & \textbf{0.40} & \textbf{0.56} & \textbf{0.56} & \textbf{0.27} & \textbf{0.88} & \textbf{0.48} & \textbf{0.56} & \textbf{0.58} & \underline{0.86} & \textbf{0.92} \\
    \midrule
    \textbf{Results@8} \\
    ENv2 & COCO & 0.28 &  & 0.22 & 0.12 & 0.27 & 0.24 & 0.30 & 0.21 & 0.28 & 0.32 & 0.13 & 0.51 & 0.47 & 0.37 & 0.50 & 0.35 & 0.34 \\
    CLIPCap & COCO  & 0.26 &  & 0.20 & 0.10 & 0.31 & 0.27 & 0.31 & 0.21 & 0.28 & 0.26 & 0.12 & 0.48 & 0.43 & 0.31 & 0.49 & 0.32 & 0.67 \\
    CLIPCap & CC  & 0.37 &  & 0.23 & 0.31 & 0.51 & 0.34 & 0.31 & 0.32 & 0.40 & 0.45 & 0.22 & 0.59 & 0.45 & 0.51 & 0.49 & 0.32 & 0.70 \\
    CLIPCap & CC+WW & 0.21 &  & 0.06 & 0.16 & 0.25 & 0.16 & \textbf{0.62} & 0.16 & 0.20 & 0.24 & 0.12 & 0.51 & 0.20 & 0.23 & 0.26 & 0.18 & 0.32 \\
    EncDec & pHHI & \underline{0.44} &  & \textbf{0.33} & \textbf{0.38} & \underline{0.55} & \underline{0.44} & \underline{0.42} & \underline{0.34} & \underline{0.54} & \underline{0.56} & \underline{0.25} & \underline{0.85} & \underline{0.49} & \underline{0.56} & \underline{0.59} & \textbf{0.94} & \underline{0.91} \\
    CLIPCap & CC+pHHI & \textbf{0.47} &  & \textbf{0.33} & \underline{0.37} & \textbf{0.90} & \textbf{0.46} & 0.41 & \textbf{0.43} & \textbf{0.60} & \textbf{0.59} & \textbf{0.29} & \textbf{0.92} & \textbf{0.50} & \textbf{0.59} & \textbf{0.61} & \underline{0.91} & \textbf{0.96} \\
    \bottomrule
  \end{tabularx}
  \vspace{0pt}
  \caption{Results on \textbf{\imsituhhi}. In addition to the average verb embedding similarity between predicted verbs and the ground truth verb, we also present mean similarities for the most common 15 verbs in \imsituhhi. Best results are in bold, and second best are underlined. For models using beam search, we report results for top 1, 5, and 8 beams.}
\label{tab:imsitu_results}
\end{table*}

%% file: 05-04-discussion.tex
\subsection{Results and Discussion}

For \datasetname{}, we report all of the metrics described above. For \imsituhhi{}, we only use the verb similarity metric since the ground truth label is a single verb. We report average similarity over all samples in \imsituhhi as well as displaying averages for the most-supported verbs. See Tables \ref{tab:wenda_results}--\ref{tab:imsitu_results} for quantitative results, and see Figure \ref{fig:results_wenda} for a visual comparison on \datasetname{}. Note that we do not include CoFormer in the table of \imsituhhi results since it was trained directly on some of these items; see the supplementary material for analysis of CoFormer on in-distribution and out-of-distribution images in \imsituhhi.

Overall we see that training on our pseudo-labels improves performance on our benchmarks. In Tables \ref{tab:wenda_results} and \ref{tab:wenda_source_results}, showing results on \datasetname{}, the best-performing model by all metrics is CLIPCap fine-tuned with our pseudo-labels. This holds across data sources, as seen in Table \ref{tab:wenda_source_results}, showing that this improvement generalizes to images beyond those originating in the Who's Waldo dataset. This model is also the best-performing on average and across a majority of verb categories on \imsituhhi as seen in Table \ref{tab:imsitu_results}. Qualitative comparison shows that the captioning models used as-is output text that is far from the ground truth HHI labels, containing many irrelevant details and not necessarily describing an interaction. This can be seen in Figure \ref{fig:results_wenda}, where the CLIPCap (CC) captions contain many hallucinated, non-factual details.

While transfer learning with pretrained CLIPCap yields the best results, we also observe that the vanilla Encoder-Decoder fine-tuned on our pseudo-labels also performs well, achieving the second-best BLEURT score on \datasetname{} and second-best verbal similarity metrics overall and across many verb categories on \imsituhhi. We infer that our pseudo-labels do impart semantic knowledge of HHI beyond simply cueing existing captioning models to the surface form of HHI labels. Nevertheless, CLIPCap fine-tuned on pHHI does generalize better across the data from all sources in \datasetname{} and to \imsituhhi which is entirely out-of-distribution for this model.

We also note that the metrics improve dramatically for both datasets when considering 5 or 8 beams. This is consistent with the fact that beam search using models fine-tuned on pHHI outputs a list of diverse candidate interactions, allowing a more directed search in the space of HHI descriptors, while beam search applied to captioning models as-is tends to produce many slight variations of the same long caption.

%% file: 06-conclusion.tex
We present a new framework for learning to understand human-human interactions in still images using weak supervision from textual captions. We demonstrate the use of knowledge distillation applied to a large language model without explicit supervision to produce pseudo-labels that can serve as targets for predicting interactions as free text. We show that training on these pseudo-labels enables HHI understanding beyond that of SOTA captioning and situation recognition models, and we provide the \datasetname{} as a new benchmark for this task.

There are various avenues for future research to extend our work. One possible direction is the incorporation of visual grounding into HHI understanding. We predict the most salient interaction in an image, which we assume to be the interaction the one that is described or suggested in its accompanying caption. It remains to localize the participants, including generalizing to group interactions where more than two participants are visible. Another important aspect that remains to be explored is the hierarchical nature of interactions. For example, the generic HHI label ``meeting'' is valid for almost every image, while ``shaking hands'' is more specific and valid for a subset of those images. Further research could extend our results to hierarchical prediction of multiple HHI labels for a single image.

Finally, we note the importance of style-content disentanglement in HHI prediction, which our work does not explicitly consider.  Scene cues in images can be important for correctly identifying HHI, as illustrated in Figure \ref{fig:teaser}, but also may be misleading. For instance, an image of soldiers in uniform is more likely to depict ``saluting'', but HHI is only valid if the image actually contains a salute. Future work on disentangling style and content shows promise for improving the robustness of HHI understanding models.

%% file: 07-ack.tex
\medskip \noindent \textbf{Acknowledgements.} We thank Ron Mokady for providing helpful feedback. This work was supported by a research gift from Meta and the Alon fellowship.

%% file: supp-1-03-coformer-imsitu.tex
\subsection{CoFormer on \imsituhhi} \label{sec:coformer_imsitu}

 The CoFormer grounded situation recognition model, whose results on \datasetname{} are reported in the main paper, was trained on the SWiG dataset, which extends the imSitu dataset with grounding information.~\cite{cho2022collaborative,pratt2020grounded,yatskar2016} Since \imsituhhi also includes some of this data, CoFormer's performance on \imsituhhi is not comparable to the out-of-distribution performance of the other models we consider. Nevertheless, we can use its performance on \imsituhhi as a rough upper bound for this task. We report its performance on all of \imsituhhi, which includes some of its training data, as well as on the intersection of \imsituhhi with imSitu's train, dev, and test sets alone. See Table \ref{tab:coformer_imsitu_results} for these metrics and a comparison to the Enc-Dec model trained on our pseudo-labels. As expected, CoFormer's performance is much higher on its own training data, and generally outperforms our model by this metric on imSitu. However, CoFormer was trained using the verb labels from imSitu, while our model, trained without supervision from manually-labelled data, is being evaluated out-of-distribution and without regard to the additional text in its predictions besides the predicted verb.

 \input{tables/supp_coformer_imsitu_results}

%% file: tables/supp_coformer_imsitu_results.tex
\begin{table}[t]
  \centering
  \setlength{\tabcolsep}{5.5pt}
  \def\arraystretch{0.95}
  \begin{tabularx}{0.75\columnwidth}{llcc}
    \toprule
    Model & Data & Eval split & sim.   \\
    \midrule
    CoFormer & SWiG & all (8021) & 0.63 \\
    EncDec & pHHI & all (8021) & 0.28 \\
    \midrule
    CoFormer & SWiG & train (4906) & 0.73 \\
    EncDec & pHHI & train (4906) & 0.34 \\
    \midrule
    CoFormer & SWiG & dev (1549) & 0.50 \\
    EncDec & pHHI & dev (1549) & 0.27 \\
    \midrule
    CoFormer & SWiG & test (1566) & 0.48 \\
    EncDec & pHHI & test (1566) & 0.28 \\
    \bottomrule
  \end{tabularx}
  \vspace{5pt}
  \caption{CoFormer results on \textbf{\imsituhhi} as described in Section \ref{sec:coformer_imsitu}, with Enc-Dec model for comparison. CoFormer was trained with supervision from the imSitu train set, while our models did not see any of these samples during training; therefore, we treat the CoFormer model performance as an upper bound for achievable verb similarity on this dataset in the out-of-distribution setting. The ``Data'' column shows the model's training data. The ``Eval split'' column gives the evaluation data split used and its size - either the entire 8,021-sample \imsituhhi subset of imSitu, or else its intersection with imSitu's train, dev, or test sets. The average verb embedding similarity is shown as ``sim.''. Note that SWiG here refers to the train set of imSitu along with grounding data. Enc-Dec model results refer to top-1 predictions.}
\label{tab:coformer_imsitu_results}
\end{table}

%% file: supp-1-04-extended-imsitu.tex
\subsection{Extended-\imsituhhi results} \label{sec:ext}

In Section \ref{sec:imsitu_details}, we described the construction of the 8,021-sample \imsituhhi dataset, a subset of the full imSitu dataset. One of its design choices was the final filtering of verbs by number of supported images, to use only those verbs with at least 100 images after filtering for human detections and semantic arguments. We now present results on an extended version of this dataset where we lower the threshold for the required number of images supporting a verb and thus keep a larger subset of imSitu.

See Table \ref{tab:extended_imsitu_results} for quantitative results. We observe that decreasing the minimum required support of verbs increases the number of unique verbs dramatically, but has a minimal impact on the verb embedding similarity metric when lowered from 100 to 50. However, lower thresholds more significantly impact the verb similarity scores. This comports with the observation that verbs with higher support values are more likely to represent HHI.

We include examples of verbs with support values at different levels to illustrate this intuition:

\textbf{Verbs with support $\geq 180$}: \emph{socializing, distributing, teaching, communicating, interviewing, lecturing, training, providing, instructing, giving, pushing, helping, asking, coaching, selling, talking, educating}

\textbf{Verbs with support $\in [100, 120]$}: \emph{imitating, offering, plunging, pitching, reassuring, autographing, clapping, ignoring, dousing, speaking, operating, wheeling, loading}

\textbf{Verbs with support $\in [50, 55]$}: \emph{repairing, chasing, drumming, applauding, breaking, eating, climbing, officiating, carting, deflecting, building, measuring}

\textbf{Verbs with support $\in [20, 25]$}: \emph{colliding, guarding, submerging, twirling, rocking, miming, clearing, calming, sowing, massaging, nuzzling, butting, tasting, waxing, clenching, knocking, scooping, stacking, vaulting, shopping}

\textbf{Verbs with support $\in [1, 2]$}: \emph{curtsying, coughing, reading, crawling, surfing, dialing, erasing, slipping, marching, frying, dripping, phoning, mopping, bulldozing, sharpening, walking, landing, boating, circling, boarding, skipping, shivering, signing, flapping, crouching, sneezing, raking, launching, protesting, piloting, unplugging, ejecting, praying, typing, stitching, watering, queuing}

\input{tables/supp_extended_imsitu_results}

%% file: tables/supp_extended_imsitu_results.tex
\begin{table}[t]
  \centering
  \setlength{\tabcolsep}{5.5pt}
  \def\arraystretch{0.95}
  \begin{tabularx}{0.99\columnwidth}{cccccc}
    \toprule
    Support & Verbs & Samples & sim@1 & sim@5 & sim@8   \\
    \midrule
    $\geq 100$ & 50 & ${\sim}8k$ & 0.28 & 0.40 & 0.44 \\
    $\geq 50$ & 98 & ${\sim}11k$ & 0.28 & 0.40 & 0.43 \\
    $\geq 20$ & 178 & ${\sim}14k$ & 0.26 & 0.38 & 0.41 \\
    $\geq 0$ & 359 & ${\sim}15k$ & 0.25 & 0.37 & 0.40 \\
    \bottomrule
  \end{tabularx}
  \vspace{5pt}
  \caption{Results of the EncDec model on extended-\imsituhhi, as described in Section \ref{sec:ext}.}
\label{tab:extended_imsitu_results}
\end{table}

%% file: supp-1-05-extractive-only.tex
\subsection{Training on syntactic parsing-based seeds} \label{sec:extonly}

To ablate the effect of our pseudo-labelling, we compare to results when training directly on syntactic parsing-based seeds. As described in the main paper, these can sometimes be extracted from Who's Waldo captions when they fit a particular syntactic pattern, specifically containing an interaction verb with arguments representing the relevant participants.

Out of the ${\sim}126k$ images from Who's Waldo that we used, only ${\sim}23k$ have captions that yield a syntactic parsing-based seed (while pseudo-labels could be assigned to all of them). Therefore in this ablation the models train on $< 20\%$ the number of images used to train the models with pseudo-labelling.

We compare results on \datasetname{} and \imsituhhi when training only on these seeds versus training on our pseudo-labels in Table \ref{tab:supp_extonly_results}. In addition to the textual similarity metrics, we include two simple measures of diversity: the number of unique interaction texts in the predictions ($n_i$) and the number of unique predicted verbs ($n_v$) across all test items. Although diversity metrics are less meaningful for comparisons to the output of captioning models used as-is, since their outputs are highly detailed, they can be used in this case since the models under comparison all output predictions of roughly the same length and level of detail. Models trained on pHHI show higher similarity to the ground truth labels as seen in the reported textual similarity metrics. In addition, we see a significant increase in diversity relative to training on syntactic parsing seeds. This suggests that the large increase in training data provided by pseudo-labelling allows models to represent a larger space of interactions, consistent with our goal in modelling the heavy tail of possible HHI. This is also illustrated in Figure 
\ref{fig:supp_diversity_examples}, which compares outputs of two models (both pretrained on CC captions)---one trained with our pseudo-labels and the other with the set of syntactic parsing-based seeds.

\input{tables/supp_extonly_results}

\input{figures/supp/supp_diversity_examples}

%% file: tables/supp_extonly_results.tex
\begin{table}[t]
  \centering
  \setlength{\tabcolsep}{2.3pt}
  \def\arraystretch{0.95}
  \begin{tabularx}{0.48\textwidth}{ccccccccc}
    \toprule
    & & \multicolumn{4}{c}{\datasetname{}} & \multicolumn{3}{c}{\imsituhhi{}} 
    \\
    \cmidrule(lr){3-6}
    \cmidrule(lr){7-9}
    Method & Data & BL & sim & $n_i$ & $n_v$ & sim & $n_i$ & $n_v$\\
    \midrule
    EncDec & pHHI & 0.38 & 0.41 & 298 & 100 & 0.28 & 1468 & 245 \\
    CLIPCap & CC+pHHI & 0.42 & 0.46 & 158 & 86 & 0.32 & 325 & 133 \\
    \midrule
    EncDec & SP & 0.33 & 0.36 & 126 & 66 & 0.24 & 216 & 82 \\
    CLIPCap & CC+SP & 0.41 & 0.44 & 123 & 78 & 0.29 & 268 & 129 \\
    \bottomrule
  \end{tabularx}
  \vspace{5pt}
  \caption{Comparison of results when training on syntactic parsing-based seeds (``SP'') versus our pseudo-labels (``pHHI''), as described in Section \ref{sec:extonly}. ``BL'' refers to BLEURT and ``sim'' refers to verb embedding similarity. On \datasetname{}, results are aggregated across data sources.}
\label{tab:supp_extonly_results}
\end{table}

%% file: figures/supp/supp_diversity_examples.tex
\begin{figure*}
  \jsubfig{\includegraphics[height=2.88cm]{figures/results/wenda/white.png}}{\begin{flushleft}\vspace{-15pt} \small{ CLIPCap \newline (CC+pHHI) \newline \vspace{-3pt} \newline CLIPCap (CC+SP)}
  \end{flushleft}} 
  \hfill
  \hfill
  \jsubfig{\includegraphics[height=3.5cm]{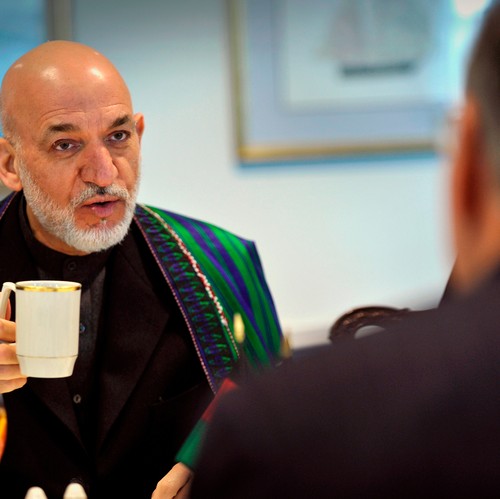}}{\vspace{-5pt}\begin{flushleft} \small{  \SNAME{} being interviewed by \SNAME{} \\ \vspace{10pt}     \SNAME{} talking with \SNAME{}} \end{flushleft}}
          \hfill
  \jsubfig{\includegraphics[height=3.5cm]{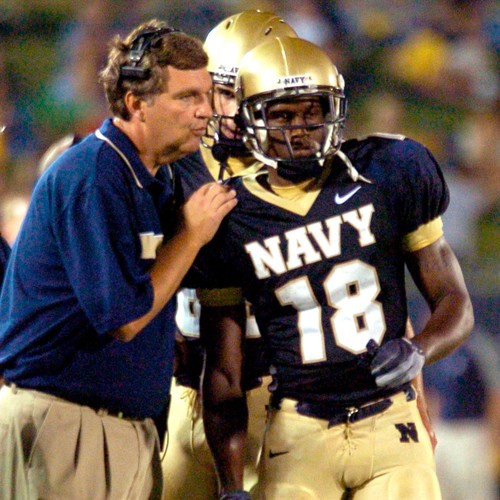}}{\vspace{-5pt}\begin{flushleft} \small{ \SNAME{} coaching \SNAME{}  \\ \vspace{22pt}  \SNAME{} talking with \SNAME{} } \end{flushleft}}
          \hfill
\jsubfig{\includegraphics[height=3.5cm]{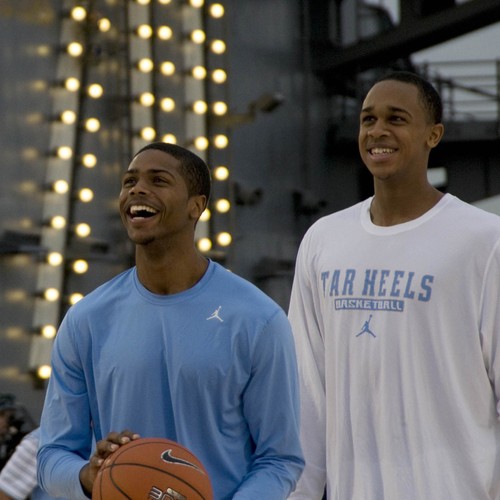}}{\vspace{-5pt}\begin{flushleft} \small{ \SNAME{} playing basketball with \SNAME{}  \\ \vspace{10pt}     \SNAME{} playing with \SNAME{} } \end{flushleft}}
          \hfill
  \jsubfig{\includegraphics[height=3.5cm]{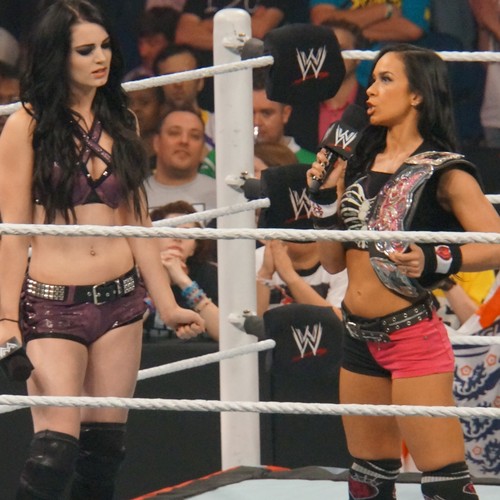}}{\vspace{-5pt}\begin{flushleft} \small{  \SNAME{} wrestling with \SNAME{}  \\ \vspace{22pt}     \SNAME{} playing with \SNAME{} } \end{flushleft}}
\vspace{0pt}
  \caption{Examples of diverse predictions on \datasetname{} from a model trained with our pseudo-labels, compared to predictions when trained on syntactic-parsing based seeds (``SP''). See Section \ref{sec:extonly} for details.
  }\label{fig:supp_diversity_examples}
\end{figure*}

%% file: supp-1-06a-neural.tex
\subsection{Additional neural metrics} \label{sec:neural}

In addition to BLEURT, we report metrics for additional neural metrics for natural language generation. Fmor easuring textual similarity between predictions and ground truth HHI labels, we provide results for BERTScore~\cite{zhang2019bertscore} and BARTScore~\cite{yuan2021bartscore}. We also measure factuality of predictions relative to ground truth captions (similar to the NLI scores reported in the main paper) using the model SummaC~\cite{laban2022summac}. Table \ref{tab:supp_neural_results} for results on captioning models, aggregated over data sources in \datasetname{}.

BERTScore uses the default pretrained checkpoint for English provided by the Hugging Face \texttt{evaluate} library\footnote{\url{https://huggingface.co/spaces/evaluate-metric/bertscore}}, and we report the output F1 score. BARTScore uses the model trained on ParaBank2 provied in the official BARTScore repository\footnote{\url{https://github.com/neulab/BARTScore}}. SummaC scores use the default checkpoint and settings for SummaC-Conv provided in its official repository\footnote{\url{https://github.com/tingofurro/summac}}. For all of these models, we replace \NAME{} with the text ``person'' as needed, just as we do for calculating BLEURT scores (see Section \ref{sec:supp_metrics})

We see the textual similarity metrics (BERTScore, BARTScore) pattern similarly to BLEURT in supporting the use of our pHHI as training data. SummaC scores are slightly higher for captioning models trained on COCO and used as-is, possibly reflecting generic text that is closer to ground truth captions though not necessarily effective at capturing HHI.

\input{tables/supp_neural_results}

%% file: tables/supp_neural_results.tex
\begin{table}[t]
  \centering
  \setlength{\tabcolsep}{5.5pt}
  \def\arraystretch{0.95}
  \begin{tabularx}{0.95\columnwidth}{ccccccc}
    \toprule
    Method & Data & & BL & BE & BA & SC \\
    \midrule
    ENv2 & COCO & & 0.27 & 0.87 & -6.25 & \textbf{0.24} \\
    CLIPCap & COCO & & 0.28 & 0.87 & -7.24 & \textbf{0.24} \\
    CLIPCap & CC & & 0.27 & 0.86 & -6.66 & 0.23 \\
    CLIPCap & CC+WW & & 0.26 & 0.85 & -5.90 & 0.22 \\
    EncDec & pHHI & & \underline{0.38} & \underline{0.92} & \underline{-3.53} & 0.22 \\
    CLIPCap & CC+pHHI & & \textbf{0.42} & \textbf{0.93} & \textbf{-3.34} & 0.22 \\
    \bottomrule
  \end{tabularx}
  \vspace{5pt}
  \caption{Comparison of BLEURT and additional neural metrics on captioning models, aggregated across data sources in \datasetname{}, as described in Section \ref{sec:neural}. Metrics shown are BLEURT (BL), BERTScore (BE), BARTScore (BA), and SummaC (SC).}
\label{tab:supp_neural_results}
\end{table}

%% file: supp-1-06b-ngram.tex
\subsection{Ngram-based metrics} \label{sec:ngram}

In this section we discuss the use of BLEURT~\cite{sellam2020bleurt} as our main textual metric rather than ngram-based metrics such as BLEU~\cite{papineni2002bleu}. Ngram-based metrics are common in text generation tasks such as machine translation, comparing predicted texts to a ground truth reference (or multiple references). They have the advantage of being simple and fast to calculate, but focus on surface forms of text rather than underlying semantics.

We provide a comparison of BLEU and BLEURT scores in Table \ref{tab:supp_ngram_results}, aggregated across data sources in \datasetname{}. We provide scores for a captioning model (CLIPCap trained on Conceptual Captions) and a model fine-tuned on our pseudo-labels. Although the latter has a higher BLEU score, its extremely low value (0.06) is due to the fact that only 72 out of 1,000 predictions achieve a nonzero BLEU score relative to the ground truth labels. Because BLEU measures ngram precision and the ground truth labels are short, it returns zero unless the prediction is a near-perfect textual match. This effectively ignores the vast majority of predictions, unlike BLEURT which \cite{sellam2020bleurt} show to have a robust correlation with human judgements of semantic similarity at the sentence level.

We additionally provide scores for the METEOR metric~\cite{banarjee2005}, which uses unigram alignment statistics and incorporates both precision and recall. It also uses stemming and synonym matching to provide some robustness relative to changes in the surface forms of semantically similar texts. Although METEOR does not ignore most predictions as does BLEU, we find that it underperforms BLEURT in capturing semantic similarity in our setting. The baselines in Table \ref{tab:supp_ngram_results} are calculated by (1) using the constant text ``\NAME meeting with \NAME'', and (2) randomizing the order of the predictions of the model fine-tuned on our pseudo-labels. Both baselines achieve a relatively high METEOR score, while BLEURT decreases significantly and approaches the BLEURT score of the plain captioning model. This suggests that METEOR is biased towards measuring surface similarity rather than underlying semantics, consistent with the findings of \cite{sellam2020bleurt} who explicitly compare METEOR and BLEURT. This can also be seen in the qualitative examples in Table \ref{tab:supp_ngram_qualit} of prediction (CLIPCap CC+pHHI) and ground truth pairs from \datasetname{} where METEOR and BLEURT differ strongly in magnitude.

We replace \NAME{} with the text ``person'' as needed to calculate these scores, just as we do for calculating BLEURT scores (see Section \ref{sec:supp_metrics}).

\input{tables/supp_ngram_results}
\input{tables/supp_ngram_qualit}

%% file: tables/supp_ngram_results.tex
\begin{table}[t]
  \centering
  \setlength{\tabcolsep}{5.5pt}
  \def\arraystretch{0.95}
  \begin{tabularx}{0.99\columnwidth}{ccccc}
    \toprule
    Method & Data & BLEU & METEOR & BLEURT \\
    \midrule
    CLIPCap & CC & 0.00 & 0.13 & 0.27 \\
    CLIPCap & CC+pHHI & 0.06 & 0.46 & 0.42 \\
    \midrule
    \multicolumn{2}{c}{baseline (constant)} & 0.00 & 0.36 & 0.29 \\
    \multicolumn{2}{c}{baseline (jumbled)} & 0.00 & 0.43 & 0.33 \\
    \bottomrule
  \end{tabularx}
  \vspace{5pt}
  \caption{Comparison of ngram-based metrics and BLEURT, aggregated across data sources in \datasetname{}, as described in Section \ref{sec:ngram}.}
\label{tab:supp_ngram_results}
\end{table}

%% file: tables/supp_ngram_qualit.tex
\begin{table*}[t]
  \centering
  \setlength{\tabcolsep}{5.5pt}
  \def\arraystretch{0.95}
  \begin{tabularx}{0.8\textwidth}{llcc}
    \toprule
    Ground truth & Prediction & METEOR & BLEURT \\
    \midrule
    \SNAME{} wrestling with \SNAME{} & \SNAME{} competing against \SNAME{} & 0.25 & 0.65 \\
    \SNAME{} giving signatures to \SNAME{} & \SNAME{} signing autographs with \SNAME{} & 0.20 & 0.57 \\
    \SNAME{} arguing with \SNAME{} & \SNAME{} driving with \SNAME{} & 0.64 & 0.28 \\
    \SNAME{} making sandcastles with \SNAME{} & \SNAME{} working with \SNAME{} & 0.77 & 0.24 \\
    \bottomrule
  \end{tabularx}
  \vspace{5pt}
  \caption{Comparison of METEOR and BLEURT scores on selected examples from \datasetname{}, as described in Section \ref{sec:ngram}. Predictions are from CLIPCap trained on Conceptual Captions and fine-tuned on our pseudo-labels.}
\label{tab:supp_ngram_qualit}
\end{table*}

%% file: supp-1-07-abl.tex
\subsection{Ablation of few-shot learning for synthetic caption generation}

In order to ablate few-shot examples used when generating synthetic captions (see Section \ref{sec:gen}), we split our synthetic caption-interaction pairs into two non-overlapping folds, train summarization models on each of these folds and
then generate pseudo-labels with each model. We fine-tune
CLIPCap+CC on these pseudo-labels and evaluate the resulting models on \datasetname{}, as shown in Table \ref{tab:supp_fewshot_abl}. The negligible differences across all metrics suggest that our method is robust to the particular (randomly selected) few-shot examples used in training the summarizer.

\input{tables/supp_fewshot_abl}

%% file: tables/supp_fewshot_abl.tex
\begin{table}[t]
  \centering
  \setlength{\tabcolsep}{5.5pt}
  \def\arraystretch{0.95}
  \begin{tabularx}{0.99\columnwidth}{llcccc}
    \toprule
    Method & Training Data & BL$\,\uparrow$ & $p_e \uparrow$ & $p_c \downarrow$ & sim$\,\uparrow$   \\
    \midrule
    CLIPCap &CC+$\text{pHHI}_1$ & 0.39 &	0.35 &	0.35 & 0.43 \\
    CLIPCap &CC+$\text{pHHI}_2$ & 0.39 &	0.35 &	0.37 & 0.42 \\
    \bottomrule
  \end{tabularx}
  \vspace{5pt}
  \caption{
  Few-shot learning ablation. $\text{pHHI}_1$ and $\text{pHHI}_2$ refer to pseudo-labels produced from summarizers trained on two non-overlapping splits of our synthetic caption-interaction data.
  }
\label{tab:supp_fewshot_abl}
\end{table}

%% file: supp-1-08-qualitative.tex
\subsection{Qualitative results} \label{sec:qual}

See \href{https://learning-interactions.github.io}{our project page} for an interactive visualization of the results of all of the considered models on the \datasetname{} 1,000-item test set and on the 8,021-item \imsituhhi dataset.

%% file: supp-2-01-cc-news-captions.tex
\subsection{Scraping additional captions from CC-News}

In order to find additional caption texts for use in our knowledge distillation process, we use the CC-News dataset as available via Hugging Face datasets\footnote{\url{https://huggingface.co/datasets/cc_news}}, containing the text of ${\sim}708k$ scraped English language news articles from 2017 through 2019~\cite{Hamborg2017}. These frequently include the text of captions accompanying images in news articles. To roughly filter for these captions, we select lines of $\leq 1,000$ characters that contain any of the following textual patterns: ``(left)'', ``(right)'', ``(center)'',  ``, left,'', ``, right,'', ``, center,'',  ``, centre,'', ``, pictured,'', ``PHOTO: '', ``Photo by'', ``Image copyright'', ``Getty '', ``AP Photo'', ``AP Image''.

In captions that we extract, we remove those patterns along with the following, so that the extracted captions will not all contain common substrings:  ``(Image ...)'', ``(Photo ...)'', ``(AP Photo ...)'',  ``(Credit ...)'',  ``[Image ...]'', ``[Featured Image ...]'', ``Getty Images'',  ``Image copyright ... Image caption'', ``Photo:'',  ``FILE PHOTO:'',  ``Image (number) of (number)''.

Finally, we discard captions that did not contain an interaction as extracted in Section \ref{sec:parsing}. This left us with 6,212 captions. Examples of such captions from CC-News include the following (patterns detected and removed are shown in red strike-through text):

\begin{itemize}
    \item Northern Ireland's Corry Evans\cancel{, left,} and Germany's Toni Kroos battle for the ball during their 2018 World Cup Group C qualifying soccer match at Windsor Park, Belfast, Thursday, Oct. 5, 2017. (Brian Lawless/PA via AP)
    \item Arizona Coyotes defenseman Luke Schenn (2) and Los Angeles Kings left winger Kyle Clifford (13) reach for the puck during the second period of an NHL hockey game in Los Angeles on Saturday, Feb. 3, 2018.\cancel{ (AP Photo/Reed Saxon)}
    \item \cancel{Image copyright }Kalpana Vaughan Wilson \cancel{Image caption }Kalpana Wilson pictured with daughter Clara shortly after giving birth
\end{itemize}

%% file: supp-2-02-parsing-interactions.tex
\subsection{Syntactic parsing-based interactions} \label{sec:parsing}

We use syntactic parsing with spaCy's \texttt{en\_core\_web\_trf} model to extract interactions from CC-News and Who's Waldo captions using the procedure described below. As described in Section \ref{sec:novel}, the parsing-based interactions from CC-News are used as seeds to generate more novel interaction texts. Then, as further described in \ref{sec:pseudo}, the interactions from Who's Waldo and novel interactions are used to generate synthetic interaction-caption pairs for use in training a summarization model.

For each caption, we search for verbs that it contains. For each verb lemma V, we consider all of its children in the syntactic parse tree. For child node X, we extract the text of X's syntactic head. If X is a preposition, we also extract the head of its complement, and for any determined noun we also extract the text of its determiner. We filter out any such X containing named entities of types \texttt{DATE}, \texttt{GPE}, \texttt{FAC}, \texttt{ORG}, \texttt{LOC}, or \texttt{TIME}, and if X contains coordinated human named entities (``\texttt{NAME} and \texttt{NAME}'') we include both of them. We mask all human name entities using the special token \NAME{}. We concatenate all of these together, including V in present continuous form, to form an extractive interaction text. Finally we filter for such texts containing at least two \texttt{NAME} entities, with at least one of them being a syntactic subject\footnote{The default entity labels in this parsing model use the label \texttt{PERSON} for human entities, but we use \texttt{NAME} for consistency with later sections.}.

Also note that since captions from Who's Waldo already have human names masked as \NAME{}, we first replaced these tokens with generic names (``Adam, Bob, ...'') before applying syntactic parsing, so that the input text would be valid English.

Among CC-News captions, 6,212 captions include such interactions. Interactions extracted from CC-News captions include the following:

\begin{itemize}
    \item 
    \begin{itemize}
        \item \textbf{CC-News caption:} Chinese President Xi Jinping (L) and First Lady Peng Liyuan bid farewell as they board their plane to depart from the Julius Nyerere International Airport in Dar es Salaam, Tanzania, March 25, 2013. REUTERS/Thomas Mukoya/File Photo
        \item \textbf{Extracted interaction:} \NAME{} and \NAME{} bidding farewell
    \end{itemize}
    \item 
    \begin{itemize}
        \item \textbf{CC-News caption:} Colombia’s Radamel Falcao jumps for the ball with England’s Harry Maguire during the round of 16 match between Colombia and England at the 2018 soccer World Cup in the Spartak Stadium, in Moscow, Russia, Tuesday, July 3, 2018.
        \item \textbf{Extracted interaction:} Colombia \NAME{} jumping for the ball with England \NAME{} during the match
    \end{itemize}
    \item 
    \begin{itemize}
        \item \textbf{CC-News caption:} Chuck Munro and Brian Alexander of Spraying Systems welcome Eric Vetters of ProCorr to their booth at NACE 2018 in Phoenix.
        \item \textbf{Extracted interaction:} \NAME{} and \NAME{} welcoming \NAME{} to their booth
    \end{itemize}
\end{itemize}

In addition, 22,637 captions from Who's Waldo include such interactions. Interactions extracted from Who's Waldo captions include the following:

\begin{itemize}
    \item 
    \begin{itemize}
        \item \textbf{Who's Waldo caption:} Chief of Naval Operations Adm. \NAME{} speaks at the Navy and Marine Corps Relief Society ball with Vice Commandant of the Marine Corps Gen. \NAME{} at the Washington Hilton.
        \item \textbf{Extracted interaction:} \NAME{} speaking at the ball with \NAME{} at the Hilton
    \end{itemize}
    \item 
    \begin{itemize}
        \item \textbf{Who's Waldo caption:} \NAME{} and \NAME{} discuss Ancestry at the Maltz Performing Arts Center
        \item \textbf{Extracted interaction:} \NAME{} and \NAME{} discussing Ancestry at the Center
    \end{itemize}
    \item 
    \begin{itemize}
        \item \textbf{Who's Waldo caption:} NASA astronaut \NAME{} (left) and Japan Aerospace Exploration Agency (JAXA) astronaut \NAME{}, both Expedition 20 flight engineers, perform a check of the Synchronized Position Hold, Engage, Reorient, Experimental Satellites (SPHERES) Beacon / Beacon Tester in the Destiny laboratory of the International Space Station.
        \item \textbf{Extracted interaction:} \NAME{} and \NAME{} performing a check in the laboratory
    \end{itemize}
\end{itemize}

Note that these extracted interactions may contain prepositional phrases. We remove prepositional phrases from results when generating synthetic interaction-caption pairs, as described in Section \ref{sec:gen}.

%% file: supp-2-03-novel-interactions.tex
\subsection{Generating novel interaction texts} \label{sec:novel}

Among the 6,212 CC-News captions with interactions, we have only 3,146 unique interaction texts as extracted by the parsing-based model described above. In order to have access to a richer set of interactions for training the subsequent summarization model, we use text generation with a large language model to generate more interactions similar to those extracted from CC-News captions with the above method, using the parsing-based interactions as seeds. We use few-shot prompting by providing 10 random newline-separated parsing-based interactions from CC-News captions as a prompt to the large language model GPT-Neo-1.3B~\cite{gpt-neo,gao2020pile} and generating until the next newline. We use nucleus sampling~\cite{holtzman2019curious} with $p=0.95$, as well as a constraint to prevent repeated trigrams. We also replace \NAME{} mask tokens with generic names (``Alex, Bailey, \ldots'') so that the input text is more natural English and thus more in distribution for the language model. We discard texts that do not pass the following filters:
\begin{itemize}
    \item Text contains ``Alex'' and ``Bailey'' in order, exactly once, and no other names.
    \item Text does not contain uppercase letters, besides in names.
    \item Text must contain a word ending in ``-ing''.
    \item Text does not end with `` the'' or `` a''.
\end{itemize}

Finally, we re-mask names with the token \NAME{}. In this way we generate ${\sim}116k$ novel interaction texts used for synthetic interaction-caption pairs as described in Section \ref{sec:gen}.

Examples of such randomly generated interaction texts include the following:

\begin{itemize}
\item \NAME{} kissing \NAME{} after a win
\item \NAME{} handing \NAME{} an autograph sheet
\item \NAME{} congratulating \NAME{} in victory
\item \NAME{} calling \NAME{} in a business suit
\item \NAME{} hugging \NAME{}
\item \NAME{} telling \NAME{} he'll have
\item \NAME{} catching a short pass from \NAME{} during a play
\item \NAME{} receiving a high five from \NAME{} in the post
\item \NAME{} giving \NAME{} congratulations for a goal during a period
\item \NAME{} telling \NAME{} that he's glad he came out to see him
\item \NAME{} as \NAME{} is being picked
\item \NAME{} shooting over \NAME{} during practice
\item \NAME{} saying to \NAME{} what he is going to do
\item \NAME{} watching \NAME{} celebrate with teammates as the ceremony began
\item \NAME{} walking with \NAME{} around the deep area

\end{itemize}

As mentioned above, these may contain prepositional phrases, which are removed later as discussed in Section \ref{sec:gen}.

%% file: supp-2-04-synthetic-generation.tex
\subsection{Synthetic interaction-caption pair generation} \label{sec:gen}

Using syntactic parsing-based caption-interaction pairs from Who's Waldo data, described in Section \ref{sec:parsing}, and novel interaction texts from CC-News, described in Section \ref{sec:novel}, we use few-shot learning to generate training data for an abstractive summarization model as follows:

For each inference iteration, we construct a few-shot prompt by selecting 10 interaction-caption pairs $(I_1, C_1), \cdots, (I_k, C_k)$ using captions from Who's Waldo and syntactic parsing-based interaction texts, and a single novel CC-News based interaction $I^*$. For each pair $(I_i, C_i)$, as well as in $I^*$, we replace \NAME{} tokens with random names using the \texttt{random-name} library\footnote{\url{https://github.com/dominictarr/random-name}} library. We then construct a prompt containing the following texts, in order and newline-separated:

\begin{itemize}
    \item For $i=1,\cdots,k$:
    \begin{itemize}
        \item ``Caption of image showing $I_i$''
        \item $C_i$
    \end{itemize}
    \item ``Caption of image showing $I^*$:''
\end{itemize}

We input this prompt to GPT-Neo-1.3B~\cite{gpt-neo,gao2020pile} and generate text until a newline is output. We generate using nucleus sampling~\cite{holtzman2019curious} with $p=0.95$, temperature $0.7$, a constraint to prevent repeated trigrams, and a maximum output length of 200 tokens.

Denote the output of generation by $C^*$. The pairs $(I^*, C^*)$ generated by this method are noisy, so we select for valid synthetic interaction-caption using the following filters:

\begin{itemize}
    \item $C^*$ must contain the same random names that were used for $I^*$ in the prompt
    \item $C^*$ must entail $I^*$ ($p_e > 0.5$), as measured by the entailment probability $p_e$ calculated by a pretrained NLI model. We use BART-large~\cite{lewis2019bart} fine-tuned on the MNLI dataset~\cite{williams2017broad} (using the \texttt{facebook/bart-large-mnli} checkpoint from Hugging Face model hub\footnote{\url{https://huggingface.co/facebook/bart-large-mnli}}).
    \item $I^*$ must contain a verb, checked using spaCy's \texttt{en\_core\_web\_trf} syntactic parsing model.
    \item $I^*$ may not contain any of the following banned substrings, which are common artifacts that do not reflect interactions: ``photo'', ``image'', ``picture'', ``in this'', ``In this''
\end{itemize}

Finally, we postprocess each $I^*$ with the following steps:

\begin{itemize}
    \item Remove prepositional phrases that do not contain \NAME{}. For example: ``\NAME{} meeting with \NAME{} at a hotel'' $\rightarrow$ ``\NAME{} meeting with \NAME{}''.
    \item Normalize subjects of verbs containing two or more people joined by ``and'', ``with'', ``\&'' and/or commas, by replacing them with ``with \NAME{}'' at the end of an interaction. For example: ``\NAME{} and \NAME{} meeting'' $\rightarrow$ ``\NAME{} meeting with \NAME{}''.
\end{itemize}

It total, we generate 62,176 synthetic interaction-caption pairs with this method. Examples of such pairs include the following:

\begin{enumerate}
    \item \textbf{Caption:} Estella, a member of the Women's Auxiliary Fire Corps, hugs Lorne, the President of the United States, at a ceremony honoring firefighters at the White House in Washington, D.C. on Sept. 30, 2012. \newline \textbf{Interaction:} \NAME{} hugging \NAME{}
    \item \textbf{Caption:} Angelia shoots the puck in the face of Gladi during a game on April 27, 2012, at the St. Louis Blues home rink in St. Paul, Minn. \newline \textbf{Interaction:} \NAME{} shooting the puck against \NAME{}
    
    \item \textbf{Caption:} Emmie receives a letter in her mailbox from Jacinthe. \newline \textbf{Interaction:} \NAME{} receiving a letter from \NAME{}
    
    \item \textbf{Caption:}The hug between Bettye and Hester is a moment of joy in the life of Hester and Bettye. It was a special moment for all of them. It is a special memory for Bettye, and it is a great moment for Hester, and that's how it should be. \newline
    \textbf{Interaction:} \NAME{} hugging \NAME{}
    
    \item \textbf{Caption:} Kippie, who attended the conference, asked Paulie to make an official statement on the issue of the military's role in the US Embassy in Timor-Leste. Paulie stated that he would not comment on the matter. \newline
    \textbf{Interaction:} \NAME{} pressuring \NAME{}
\end{enumerate}

Note that although the interaction often contains the same verb as the accompanying caption, it may also contain a verb based on non-verbal cues (``hugging'' in example 4 above, with the noun ``hug'' in the caption) or even based on the general meaning of the synthetic caption (``pressuring'' in example 5 above).

%% file: supp-2-05-pseudo-labels.tex
\subsection{Pseudo-label generation} \label{sec:pseudo}

Using the synthetic interaction-caption pairs (I, C) described and illustrated in Section \ref{sec:gen}, we fine-tune a pretrained T5 model~\cite{raffel2020exploring} using the ``summarize:'' task prefix on these pairs, using each I as the target. We use \texttt{T5-base} and fine-tune for 3 epochs with batch size 8, initial learning rate $5e-5$ with linear schedule, AdamW optimizer with $(\beta_1, \beta_2) = (0.9, 0.999)$, and maximum gradient norm of $1.0$, and otherwise default hyperparameter settings as defined in the Hugging Face summarization model training script.\footnote{As of v4.18.0, script available at \url{https://github.com/huggingface/transformers/blob/31ec2cb2badfbdd4c1ac9c6c9b8a74e974984206/examples/pytorch/summarization/run_summarization.py}}

After fine-tuning, we apply this model to each caption in the Who's Waldo dataset corresponding to samples with $\geq 2$ facial detections, as provided in the dataset, to create pseudo-labels. We filter these to only keep those pseudo-labels beginning with \NAME{}, followed by a present progressive verb (``-ing''), followed by more text containing exactly one additional \NAME{}. We filter out examples containing any of the banned substrings ``photo'', ``image'', or ``picture'' since these often are artifacts that do not reflect interactions. 

Finally, in order to avoid data leakage with the test set, we remove any samples with captions identical to those in the test set, or with identical date-time metadata fields (since these often are images taken from the same event).

In total, this procedure yielded 126,696 pseudo-labels for Who's Waldo, including 1,263 unique verbs, and 16,136 unique interactions. %

%% file: supp-2-06-pseudo-labels-examples.tex
Examples of such pseudo-labels created from Who's Waldo captions include the following:
\newline

\noindent \textbf{Caption:} The Assistant Commandant of the Marine Corps, Gen. \NAME{}, \NAME{}, left, poses for a photo with Master Sgt. \NAME{} during the U.S. Marine Corps Command, Control, Communications and Computers (C4) annual awards dinner in Arlington, Va., April 17, 2014. The awards presented included the Gen. \NAME{} for outstanding communications leadership, the James Hamilton Information Technology Management Civilian Marine of the Year Award, the Pfc. Herbert A Littleton Non-Commissioned Officer Trophy for operational communications excellence, and the Lt. Col. \NAME{} Memorial Unit Award.
\newline \textbf{Pseudo-label:} \NAME{} posing with \NAME{}
\newline

\noindent \textbf{Caption:} \NAME{} and \NAME{} at Governor \NAME{} annual address in February 2016
\newline \textbf{Pseudo-label:} \NAME{} standing next to \NAME{}
\newline

\noindent \textbf{Caption:} With Italian Prime Minister \NAME{}.
\newline \textbf{Pseudo-label:} \NAME{} talking with \NAME{}
\newline

\noindent \textbf{Caption:} \NAME{} at the Gothenburg Book Fair 2014.
\newline \textbf{Pseudo-label:} \NAME{} standing with \NAME{}
\newline

\noindent \textbf{Caption:} Commemoration of 150th birth anniversary of \NAME{}, organized by the Ministry of Culture, Government of India.
\newline \textbf{Pseudo-label:} \NAME{} congratulating \NAME{}
\newline

\noindent \textbf{Caption:} General \NAME{}, Air Force Chief of Staff, addresses the 347th Wing personnel. Senator \NAME{} is standing next to the general.
\newline \textbf{Pseudo-label:} \NAME{} standing next to \NAME{}
\newline

\noindent \textbf{Caption:} Luge World Cup Men 2017/18 in Altenberg: Flower Ceremony – \NAME{}, \NAME{}, \NAME{}
\newline \textbf{Pseudo-label:} \NAME{} congratulating \NAME{}
\newline

\noindent \textbf{Caption:} US Reality TV Star And Fashion Expert \NAME{} in Sydney, by \NAME{} 'How Do I Look' was the topic of conversation at King's Cross Barrio Chino tonight. US reality television star \NAME{} and host of the 'How Do I Look' show was the main attraction. The red carpet came out as \NAME{} and a few familiar Sydney faces did their walks and poses.
\newline \textbf{Pseudo-label:} \NAME{} talking to \NAME{}
\newline

\noindent \textbf{Caption:} Crown \NAME{} and \NAME{} of Sweden during the inauguration of the Northern Link in Stockholm November 30, 2014.
\newline \textbf{Pseudo-label:} \NAME{} standing next to \NAME{}
\newline

\noindent \textbf{Caption:} \NAME{}, french politician, Brive la Gaillarde book fair, France, 2010 11 06
\newline \textbf{Pseudo-label:} \NAME{} attending \NAME{}'s book fair
\newline

\noindent \textbf{Caption:} \NAME{} during 2013 World Championships in Athletics in Moscow.
\newline \textbf{Pseudo-label:} \NAME{} standing with \NAME{}
\newline

\noindent \textbf{Caption:} \NAME{} shakes hands with Vice President \NAME{} shortly after becoming a U.S. citizen during a naturalization ceremony on Camp Victory in Baghdad, July 4, 2010. \NAME{}, assigned to the 82nd Airborne Division's 307th Brigade Support Battalion, 1st Advise and Assist Brigade, is originally from Colombia.
\newline \textbf{Pseudo-label:} \NAME{} shaking hands with \NAME{}
\newline

\noindent \textbf{Caption:} A bit of 'Underbelly' blurb that we got hold of (thanks \NAME{} - author of Razor) reads...Back in the day the East Village was called 'The Tradesman's Arms', a bloodhouse with sawdust on the floor to soak up the spit and vomit, hard stools at the bar and a dozen cheap wooden tables with chairs scattered around\&quot;. The cast of Underbelly Razor and special guests partied into the night celebrating the Underbelly Razor Uncut DVD release at the very same place that crime queens \NAME{}, \NAME{}, along with \NAME{} frequented back in their heyday. Strutting the blood red carpet was all of the Razor cast, including \NAME{}, better known now as our vice queen \NAME{}, \NAME{} who played \NAME{}, \NAME{} (\NAME{}), \NAME{}, better recognised as the \NAME{}, \NAME{}, aka the suave \NAME{}' \NAME{} and \NAME{}, who we know as \NAME{}. \NAME{} tells us of the former glory days of 'The Arms', recounted from the many interviews he conducted, compiling the book, \NAME{}. The red carpet event brought out the inner gangster in a few of us with \NAME{} stating she would consider more 'Underbelly Razor' type roles under the right circumstances, \NAME{} telling us to watch out for his uncut and fight scenes, and \NAME{} saying he was a \&quot;fashionable gangster\&quot;.
\newline \textbf{Pseudo-label:} \NAME{} hitting the red carpet with \NAME{}
\newline

\noindent \textbf{Caption:} \NAME{} and wife \NAME{}
\newline \textbf{Pseudo-label:} \NAME{} sitting with \NAME{}
\newline

\noindent \textbf{Caption:} \NAME{} at 2017 European Athletics U23 Championships
\newline \textbf{Pseudo-label:} \NAME{} standing with \NAME{}
\newline

\noindent \textbf{Caption:} \NAME{}, coach of the french feminine ski-jumping team 2010
\newline \textbf{Pseudo-label:} \NAME{} coaching \NAME{}
\newline

\noindent \textbf{Caption:} \NAME{} on the red carpet for 'Gods of Egypt' in New York City on February 24, 2016.
\newline \textbf{Pseudo-label:} \NAME{} standing with \NAME{}
\newline

\noindent \textbf{Caption:} SEOUL (July 6, 2009) Chief of Naval Operations (CNO) Adm. \NAME{} receives the National Security Merit Tongil Medal for his outstanding and meritorious service rendered to the Republic of Korea. \NAME{} is on an official visit to the U.S. 7th Fleet area of responsibility to strengthen global maritime partnerships.
\newline \textbf{Pseudo-label:} \NAME{} receiving \NAME{}'s award
\newline

\noindent \textbf{Caption:} \NAME{}, a retired United States Marine Lieutenant Colonel, and administrator at the State University of New York’s Maritime College, being promoted to two-star general in New York’s Military Forces.
\newline \textbf{Pseudo-label:} \NAME{} being promoted by \NAME{}
\newline

\noindent \textbf{Caption:} Pabradė, Lithuania -- Maj. Gen. \NAME{}, Pennsylvania’s adjutant general, shakes hands with Maj. Gen. \NAME{} in an APC 113 used by the Lithuanian Army while preparing to tour the training grounds. \NAME{} visited the exercise Amber Hope 2011 June 22 while conducting his first trip to Lithuania as Pennsylvania’s adjutant general.
\newline \textbf{Pseudo-label:} \NAME{} shaking hands with \NAME{}

%% file: supp-2-07-imsitu.tex
\subsection{\imsituhhi details} \label{sec:imsitu_details}

We form \imsituhhi, an 8,021-sample subset of the imSitu dataset~\cite{yatskar2016}, as described here.

Because we only use this data to evaluate our models, and in order to have a sufficiently large sample size in the final subset, we use all data from imSitu dataset (train, validation and test set combined together). In total this includes 126,102 samples. Using person detections from a pretrained YoloV5 model (\texttt{ultralytics/yolov5} checkpoint\footnote{\url{https://hub.docker.com/r/ultralytics/yolov5}}, pretrained on MS COCO)\cite{glenn_jocher_2021_5563715}, we discard samples whose images have less than two person detections. We also filter using the semantic frame data from imSitu, to select for samples with at least two human participants. Since arguments are not directly labelled as human or non-human, we use NLI-based filtering to select for human arguments. There are 146,347 unique argument texts in imSitu. For each such arugment A, we apply a pretrained NLI model (BART-large finetuned on MNLI, as described in Section \ref{sec:gen}) to the following pair of texts:

\begin{itemize}
    \item \textbf{Premise:} This is a \emph{A}.
    \item \textbf{Hypothesis:} This is a human.
\end{itemize}

The model returns an entailment probability $p_e$ for each such text pair, and we classify \emph{A} as a human participant if $p_e > 0.5$. We remove all samples containing less than two arguments that are classified as human.

13,560 of the unique argument texts are classified as human, including the following examples:

\begin{itemize}
    \item alpha
    \item desk sergeant
    \item Alfred the Great
    \item chief justice
    \item Gregory Pincus
    \item Pablo Neruda
    \item Spanish people
    \item abidance
    \item friend
    \item Cline
\end{itemize}

Examples of the remaining argument texts not classified as human include the following:

\begin{itemize}
    \item sugar beet
    \item barouche
    \item water development
    \item St. John's
    \item stopper
    \item horsehair
    \item stripe
    \item advocator
    \item readjustment
    \item flamingo plant
\end{itemize}

It can be seen that the arguments have a very heavy-tailed distribution, with many rare or highly specific texts, and the NLI filtering contains noise. However we find this filtering to be a useful heuristic in addition to other forms of filtering.

We filter out samples containing the following verbs with negative or inappropriate connotations: \emph{ailing, apprehending, arresting, attacking, bandaging, begging, biting, bothering, brawling, burning, clawing, complaining, confronting, crying, destroying, detaining, disciplining, dissecting, exterminating, frisking, frowning, gambling, grieving, grimacing, handcuffing, hanging, hitting, hunting, interrogating, misbehaving, mourning, panhandling, peeing, pinching, poking, pooing, pouting, punching, restraining, scolding, shooting, slapping, spanking, spearing, spying, stinging, striking, stripping, subduing, urinating, weeping, whipping}

After these filtering criteria, we are left with 15,207 samples. These samples include 359 out of the 504 unique verbs found in imSitu. The number of images supporting each verb gives an estimate of the likelihood of the given verb to describe a scenario with multiple human participants and thus gives us an estimate of its affinity to human-human interactions (HHI).

The verbs with the highest support are ``socializing'' (270 images), ``distributing'' (261 images), ``teaching'' (252 images), ``communicating'' (251 images), and ``interviewing'' (244 images). Among the least-supported verbs, which have only a single image as support,  are ``slipping'', ``skipping'', ``boarding'', ``reading'', and ``erasing''.

Finally, to select for verbs that represent HHI, only use samples with verbs that are supported by at least 100 images. This leaves us with the 8,021 \imsituhhi dataset. This contains the following 50 verbs:

\begin{itemize}
\item socializing (270 images)
\item distributing (261 images)
\item teaching (252 images)
\item communicating (251 images)
\item interviewing (244 images)
\item lecturing (241 images)
\item training (228 images)
\item providing (223 images)
\item instructing (217 images)
\item giving (213 images)
\item pushing (201 images)
\item helping (200 images)
\item asking (195 images)
\item coaching (192 images)
\item selling (185 images)
\item talking (185 images)
\item educating (183 images)
\item buying (170 images)
\item filming (161 images)
\item assembling (157 images)
\item encouraging (157 images)
\item serving (156 images)
\item dragging (155 images)
\item baptizing (153 images)
\item carrying (150 images)
\item flinging (149 images)
\item unloading (149 images)
\item crowning (145 images)
\item patting (138 images)
\item examining (132 images)
\item nagging (131 images)
\item tickling (131 images)
\item admiring (129 images)
\item shaking (123 images)
\item pinning (122 images)
\item videotaping (122 images)
\item arranging (121 images)
\item imitating (119 images)
\item offering (116 images)
\item plunging (116 images)
\item pitching (115 images)
\item reassuring (114 images)
\item autographing (112 images)
\item ignoring (109 images)
\item clapping (109 images)
\item dousing (107 images)
\item speaking (104 images)
\item operating (103 images)
\item wheeling (103 images)
\item loading (102 images)
\end{itemize}

%% file: supp-2-08-training-details.tex
\subsection{Training details} \label{sec:train}

For training CLIPCap~\cite{mokady2021clipcap}, we use checkpoints for the MLP mapping CLIPCap variant with fine-tuned GPT2 decoder, trained on Conceptual Captions.\footnote{Available at \url{https://github.com/rmokady/CLIP_prefix_caption}.}

For the Enc-Dec model, we initialize the CLIP encoder with checkpoint \texttt{vit-base-patch32} and the GPT2 decoder with checkpoint \texttt{gpt2} (base), as available in the Hugging Face transformers library.

We trained all models with batch size $16$, AdamW optimizer with learning rate $1e-5$ and $(\beta_1, \beta_2) = (0.9, 0.999)$, and weight decay $0.1$. For pretrained CLIPCap fine-tuned on our pseudo-labels, we trained for two epochs, CLIPCap trained on entire Who's Waldo captions was trained for three epochs, and the simple Enc-Dec model was trained for 17 epochs.

For models fine-tuned on our pseudo-labels, we use sample weights during training. In particular, we multiply the loss for samples with label $L$ by $c(L)^{-1/4}$, where $c(L)$ is the count of occurrences of label $L$ in our training data. In order to prevent overfitting to repeated captions in training data, we also use a multiplier of $c(C)^{-1}$ applied to training samples with caption $C$, where $c(C)$ gives the number of times caption $C$ occurs verbatim in the training data. (See Section \ref{sec:pseudo} for details on how we filter out samples with captions that are repeated in the test set.)

%% file: supp-2-09-baselines.tex
\subsection{Baseline model details} \label{sec:baselines}

As in \ref{sec:train}, pretrained CLIPCap baselines use the MLP mapping variant with fine-tuned GPT2 decoder; in this case, using both the COCO and Conceptual Captions checkpoints. For ExpansionNetV2~\cite{hu2022expansionnet}, we initialize with the weights of the ensemble model pretrained on COCO (\texttt{rf\_model.pth})\footnote{Available at \url{https://github.com/jchenghu/expansionnet_v2}.}. For CoFormer, we use the publicly available pretrained checkpoint for inference\footnote{Available at \url{https://github.com/jhcho99/CoFormer}.}.

%% file: supp-2-10-metrics.tex
\subsection{Metric calculation details} \label{sec:supp_metrics}

All reported BLEURT metrics use the BLEURT-20 checkpoint which more accurately predicts semantic similarity than the original BLEURT model~\cite{pu2021learning}. For all BLEURT calculations involving texts containing \NAME{} slots in either the predicted or ground truth text, we replace \NAME{} with the text ``person'' so that the texts are in distribution for BLEURT.

NLI metrics ($p_e, p_c$) use BART-large~\cite{lewis2019bart} fine-tuned on the MNLI dataset~\cite{williams2017broad} (using the \texttt{facebook/bart-large-mnli} checkpoint from Hugging Face model hub). To calculate these metrics, \NAME{} slots in texts are filled with an underscore character (``\_'').

Verb similarity scores use GloVe~\cite{pennington2014glove} word embeddings, specifically the \texttt{glove-wiki-gigaword-200} model available via Gensim. For models trained on our pseudo-labels, the model typically outputs the verb as the first word token, so we could use it for this metric directly. For captioning models not trained on our pseudo-labels, we virst extract a verb from their outputs for this metric using spaCy's \texttt{en\_core\_web\_trf} model. We find the first verb lemma in the given text and convert it to present continuous form (``-ing''). For texts not containing a verb, we use the zero vector as their verb embedding.

%% file: supp-2-11-coformer-eval.tex
\subsection{CoFormer evaluation details} \label{sec:coformer-eval}

Since the CoFormer baseline model does not output free text, we elaborate here on the evaluation method used to compare it to the other methods under consideration.

For all tasks, we evaluate CoFormer by using its predicted verb, discarding semantic frame and grounding predictions. This is because these semantic arguments do not directly map to the text of a human-human interaction string, so we cannot directly compare them using text-based metrics.

The results for CoFormer on \datasetname{} reported in the main paper are calculated by inserting its predicted verbs into a text prompt and treat this as the predicted interaction. We use two different prompt templates for evaluation:

\begin{itemize}
    \item $P_1$: ``\_ Ving \_'', where V denotes the given verb. This is most appropriate for transitive verbs (``\_ greeting \_'').
    \item $P_2$: ``\_ Ving with \_'', where V denotes the given verb. This is most appropriate for intransitive verbs (``\_ dancing with \_'').
\end{itemize}

Because $P_1$ or $P_2$ may be more appropriate depending on the verb, the reported metrics are aggregated by using the best (maximum or minimum, depending on the metric) score among both prompt templates for each sample.

We also note that we are discarding predicted semantic frame arguments from CoFormer's predictions that could be important to understanding the depicted interaction. However, they do not map directly to a single interaction string. Our approach has the advantage of directly inserting additional context into the predicted string using valid English syntax.

%% file: supp-3-img-attrib.tex
\section{Image Attribution}

\begin{itemize}
    \item \href{https://farm4.staticflickr.com/3608/3449266070_e4d0577481_z.jpg}{COCO val2014, ID 503278} / \href{https://creativecommons.org/licenses/by-nc-nd/2.0/}{CC BY-NC-ND 2.0}
    \item \href{https://farm4.staticflickr.com/3220/2734798500_b7bdf687d0_z.jpg}{COCO val2014, ID 369122} / \href{https://creativecommons.org/licenses/by-nc-nd/2.0/}{CC BY-NC-ND 2.0}
    \item \href{https://commons.wikimedia.org/wiki/File:US_Navy_100719-N-9818V-572_Hospital_Corpsman_1st_Class_Ingrid_Cortez,_U.S._Fleet_Forces_Sea_Sailor_of_the_Year,_gets_at_high_five_from_Master_Chief_Petty_Officer_of_the_Navy_(MCPON)_Rick_West.jpg}{Photo} by Jennifer A. Villalovos / Public domain
    \item \href{https://commons.wikimedia.org/wiki/File:Leandre_Gramss_double_double_bass_14.jpg}{Leandre Gramss double double bass 14} by \href{https://commons.wikimedia.org/wiki/User:Schorle}{Schorle} / \href{https://creativecommons.org/licenses/by-sa/4.0/deed.en}{CC BY-SA 4.0}
    \item \href{https://commons.wikimedia.org/wiki/File:2017_Ski_Tour_Canada_Quebec_city_17.jpg}{2017 Ski Tour Canada Quebec city 17} by \href{https://commons.wikimedia.org/wiki/User:Cephas}{Cephas} / \href{https://creativecommons.org/licenses/by-sa/4.0/deed.en}{CC BY-SA 4.0}
    \item \href{https://commons.wikimedia.org/wiki/File:UWS_Giants_vs._Eastlake_NEAFL_round_17,_2015_159.jpg}{UWS Giants vs. Eastlake NEAFL round 17, 2015 159} by \href{https://www.flickr.com/photos/amymergard/}{Amy Mergard} / \href{https://creativecommons.org/licenses/by/2.0/deed.en}{CC BY 2.0}
    \item \href{https://commons.wikimedia.org/wiki/File:Gansler_swearing_in.jpg}{Gansler swearing in} by \href{https://www.flickr.com/photos/28436308@N06}{Doug Gansler} / \href{https://creativecommons.org/licenses/by/2.0/deed.en}{CC BY 2.0}
    \item \href{https://commons.wikimedia.org/wiki/File:20091112_Freddie_Barnes_huddling.jpg}{20091112 Freddie Barnes huddling} by \href{https://www.flickr.com/photos/photoben27/}{PhotoBen27} / \href{https://creativecommons.org/licenses/by/2.0/deed.en}{CC BY 2.0} 
    \item \href{https://commons.wikimedia.org/?curid=33538394}{Enrique and Maja in Toronto 2014 02} by \href{https://commons.wikimedia.org/wiki/User:001Jrm}{001Jrm} / \href{https://creativecommons.org/licenses/by-sa/3.0/deed.en}{CC BY-SA 3.0}
    \item \href{https://commons.wikimedia.org/?curid=23149607}{USMC-051115-M-9876R-032} by \href{https://commons.wikimedia.org/wiki/User:Slick-o-bot}{Slick-o-bot} / Public domain
    \item \href{https://commons.wikimedia.org/?curid=25291860}{Photo} by \href{https://commons.wikimedia.org/wiki/Category:Photographs_by_Glenn_Fawcett}{Glenn Fawcett} / Public domain
    \item \href{https://commons.wikimedia.org/?curid=8227973}{Photo} by Damon J. Moritz / Public domain
    \item \href{https://commons.wikimedia.org/?curid=17312179}{Photo} by Karolina A. Martinez / Public domain
    \item \href{https://commons.wikimedia.org/?curid=32234221}{AJ Challenges Paige} by \href{https://www.flickr.com/photos/miguel_discart/}{Miguel Discart} / \href{https://creativecommons.org/licenses/by-sa/2.0/deed.en}{CC BY-SA 2.0}
\end{itemize}